\documentclass[conference]{IEEEtran}
\IEEEoverridecommandlockouts
\usepackage{cite}
\usepackage{amsmath,amssymb,amsfonts}
\usepackage{graphicx}
\usepackage{textcomp}
\usepackage{xcolor}
\def\BibTeX{{\rm B\kern-.05em{\sc i\kern-.025em b}\kern-.08em
    T\kern-.1667em\lower.7ex\hbox{E}\kern-.125emX}}

\usepackage{hyperref}
\usepackage{amsthm}

\usepackage{algorithm}
\usepackage{algpseudocode}
\usepackage{tabularx}
\usepackage{enumitem}
\usepackage{multirow}
\usepackage{wrapfig}
\usepackage{bm}
\usepackage{booktabs}

\usepackage{siunitx}
\newcolumntype{d}{S[table-format=3.2(3), separate-uncertainty]}

\makeatletter
\def\@opargbegintheorem#1#2#3{\trivlist
   \item[]{\bfseries #1\ #2\ (#3)} \itshape}
\makeatother

\makeatletter
\newcommand{\multiline}[1]{%
  \begin{tabularx}{\dimexpr\linewidth-\ALG@thistlm}[t]{@{}X@{}}
    #1
  \end{tabularx}
}
\makeatother 

\newcommand{\sysname}{MADOD}

\begin{document}

\title{\sysname{}: Generalizing OOD Detection to Unseen Domains via G-Invariance Meta-Learning}

\author{
\IEEEauthorblockN{Haoliang Wang$^1$, Chen Zhao$^2$, Feng Chen$^1$}
\IEEEauthorblockA{
$^1$\textit{Department of Computer Science, The University of Texas at Dallas, Richardson, Texas, USA} \\
$^2$\textit{Department of Computer Science, Baylor University, Waco, Texas, USA}\\
% \textit{The University of Texas at Dallas}, Richardson, USA \\
\{haoliang.wang, feng.chen\}@utdallas.edu, chen\_zhao@baylor.edu}
% \and
% \IEEEauthorblockN{Chen Zhao}
% \IEEEauthorblockA{\textit{Department of Computer Science} \\
% \textit{Baylor University}, Waco, USA \\
% chen\_zhao@baylor.edu}
% \and
% \IEEEauthorblockN{Feng Chen}
% \IEEEauthorblockA{\textit{Department of Computer Science} \\
% \textit{The University of Texas at Dallas}, Richardson, USA \\
% feng.chen@utdallas.edu}
}

\maketitle

\begin{abstract}
    Real-world machine learning applications often face simultaneous covariate and semantic shifts, challenging traditional domain generalization and out-of-distribution (OOD) detection methods. We introduce Meta-learned Across Domain Out-of-distribution Detection (\sysname{}), a novel framework designed to address both shifts concurrently. \sysname{} leverages meta-learning and G-invariance to enhance model generalizability and OOD detection in unseen domains. Our key innovation lies in task construction: we randomly designate in-distribution classes as pseudo-OODs within each meta-learning task, simulating OOD scenarios using existing data. This approach, combined with energy-based regularization, enables the learning of robust, domain-invariant features while calibrating decision boundaries for effective OOD detection. Operating in a test domain-agnostic setting, \sysname{} eliminates the need for adaptation during inference, making it suitable for scenarios where test data is unavailable. Extensive experiments on real-world and synthetic datasets demonstrate \sysname{}'s superior performance in semantic OOD detection across unseen domains, achieving an AUPR improvement of 8.48\% to 20.81\%, while maintaining competitive in-distribution classification accuracy, representing a significant advancement in handling both covariate and semantic shifts. The implementation of our framework is available at \href{https://github.com/haoliangwang86/MADOD}{https://github.com/haoliangwang86/MADOD}.

\end{abstract}


\section{Introduction}
    Semantic out-of-distribution (OOD) detection in shifted domains remains a critical challenge in machine learning, particularly in real-world applications \cite{lee2017training,grosse2017statistical,ruff2018deep,he2024gdda,lin2023adaptation,zhao2023open,wang2022layer,zhao2021fairness,zhao2022adaptive,zhao2023towards} where systems encounter both covariate shifts (changes in distributions of data features) and semantic shifts (emergence of new classes) simultaneously. Two main lines of research have emerged to address these challenges: domain generalization (DG) \cite{zhou2022domain, wang2022generalizing} for handling covariate shifts, and OOD detection \cite{yang2024generalized} for managing semantic shifts.
Figure 1 illustrates the dual challenge of covariate and semantic shifts. Domain generalization addresses covariate shift, exemplified by the transition from ``Photo", ``Art", and ``Cartoon" domains to a ``Sketch" domain, while maintaining consistent in-distribution (ID) classes such as ``Dog" and ``Elephant". OOD detection tackles semantic shift, which introduces entirely new classes like ``House" and ``Person" in the unseen target domain. While these approaches often address these shifts in isolation, the complexity of real-world scenarios demands more robust and adaptable solutions that can handle both types of shifts concurrently \cite{shu2021open,wang2023generalizable,bai2023feed}. 

\begin{figure}[!t]
    \centering
    \includegraphics[width=1\linewidth]{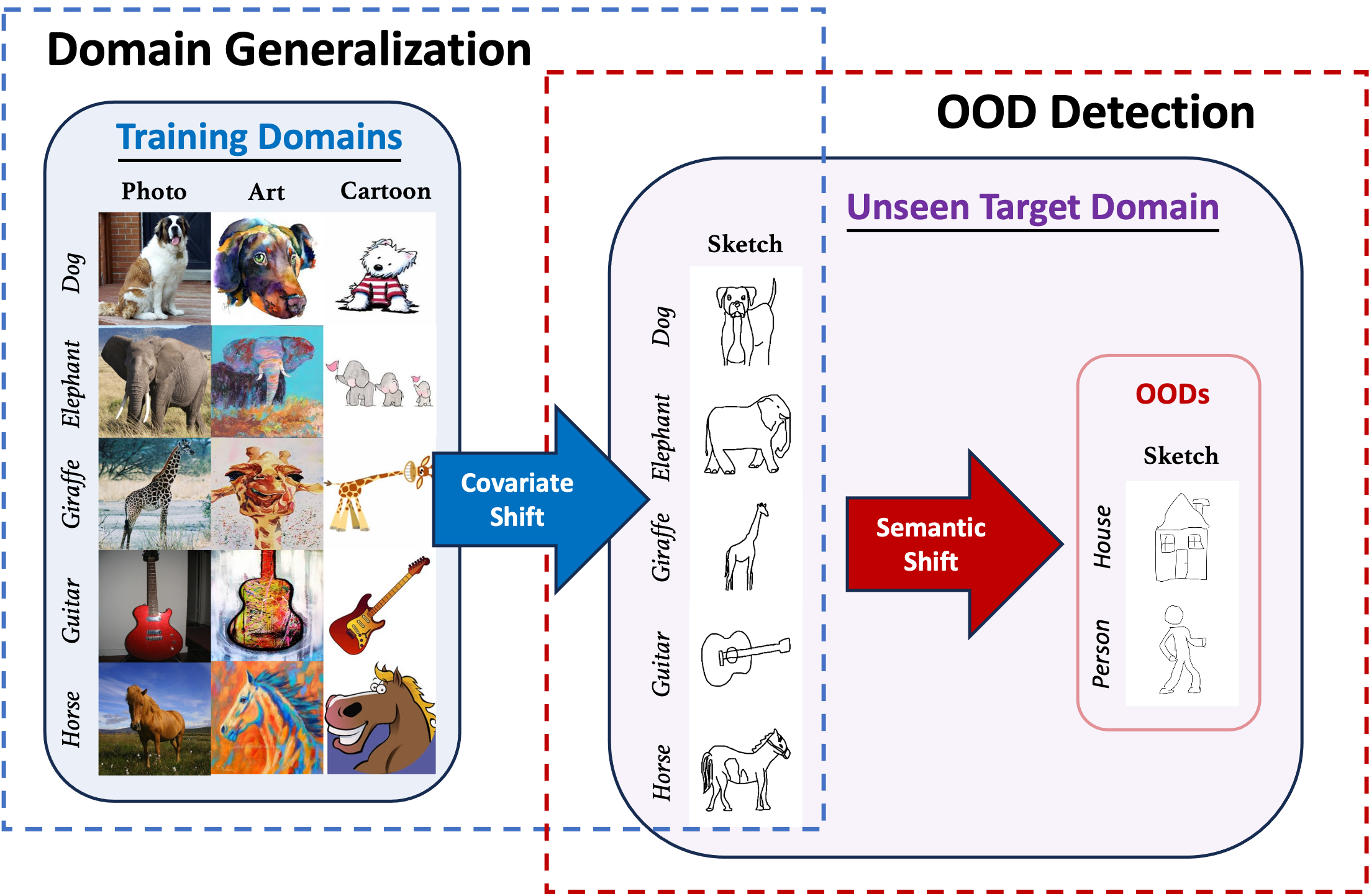}
    \caption{Problem setting for ``Semantic OOD Detection in Unseen Domains", which involves both covariate shift (testing on an unseen target domain) and semantic shift (target domain contains OOD samples with novel semantics).}
    \label{fig:problem}
\end{figure}

The problem of semantic OOD detection in unseen domains, also known as open-set domain generalization, remains relatively unexplored \cite{katsumata2021open, bai2023feed, shu2021open, wang2023generalizable, noguchi2023simple}. \textit{Katsumata et al}.~\cite{katsumata2021open} propose using metric learning to maximize the separation between known and unknown classes; however, this method relies heavily on a substantial amount of OOD training samples that may not be available in practice. Similarly, SCONE~\cite{bai2023feed} enforces an explicit energy gap between known and unknown classes using a large, unlabeled wild dataset that includes both covariate and semantic OODs, which can be challenging to obtain in practice. Other approaches, such as DAML~\cite{shu2021open}, MEDIC~\cite{wang2023generalizable}, and the method proposed by \textit{Noguchi et al}.~\cite{noguchi2023simple}, employ meta-learning strategies to expose the model to a diverse range of augmented domains or calibrate classification decision boundaries. However, these methods either require training a separate network branch for each domain or multiple one-vs-all binary classifiers, limiting their application and making them not directly compatible with state-of-the-art OOD detection techniques for improved OOD detection performance, as demonstrated in our empirical results.

Given that domain generalization (DG) is often treated as a preprocessing step by these related methods to achieve domain-invariant features, this work investigates the potential limitations of current DG techniques for this purpose. Drawing inspiration from prior studies \cite{robey2021model, huang2018multimodal, zhao2023towards, zhao2021fairness, zhao2022adaptive}, we examine a commonly adopted DG setting where inter-domain variation is driven solely by covariate shifts due to the principle of $G$-invariance, which enforces domain-invariance in learned model embeddings. We investigate how the $G$-invariance principle can be integrated into a meta-learning framework to enhance the model's generalizability further and improve semantic OOD detection in unseen domains by exposing it to a wide variety of training distributions.

Specifically, our method, \sysname{}, employs a novel task construction approach that randomly designates certain in-distribution classes as pseudo-OODs within each meta-learning task. This strategy, combined with an energy-based regularization term \cite{liu2020energy}, allows \sysname{} to learn robust, domain-invariant features, satisfying the G-invariance principle, while simultaneously calibrating its decision boundary for effective OOD detection. By integrating these components within a meta-learning framework, \sysname{} can adapt to a wide range of domain shifts and novel class distributions.

\sysname{} operates in a fully test domain-agnostic setting, setting it apart from existing meta-learning approaches and making it more applicable to real-world scenarios. Unlike traditional meta-learning frameworks that require an adaptation phase using a few test domain samples, \sysname{} assumes no access to test domain data at any point and does not rely on relative labels in task construction. Instead, we leverage the fact that our test domains share the same ID class set as the training domains, allowing us to create diverse tasks that simulate OOD scenarios using existing ID classes. This approach eliminates the need for domain adaptation during inference, making our method more suitable for real-world situations where test data may be unavailable or inaccessible.

Our main contributions are as follows:

\begin{itemize}
    \item We propose \sysname{}, a unified meta-learning framework that addresses both covariate and semantic shifts. By integrating domain generalization with OOD detection through a nested optimization strategy and $G$-invariance regularization, \sysname{} optimizes for generalization to unseen domains without requiring test data during training, ensuring effectiveness in real-world environments with unavailable domain-specific test data.

    \item We introduce an innovative task construction method that randomly selects ID classes as pseudo-OODs for each task, coupled with an energy-based regularization term. This approach enhances task diversity, simulates OOD scenarios using existing data, promotes sensitivity to distributional changes, and fine-tunes the model's decision boundary, ensuring robustness and adaptability to various OOD detection algorithms.

    \item Through extensive experiments on real-world and synthetic datasets, we demonstrate \sysname{}'s superior performance in semantic OOD detection across unseen domains, achieving an AUPR improvement of 8.48\% to 20.81\%, while maintaining competitive ID classification accuracy, validating its effectiveness in scenarios with both covariate and semantic shifts.
\end{itemize}

% By addressing the limitations of previous approaches and introducing these novel techniques, \sysname{} represents a significant advancement in the field of semantic OOD detection in unseen domains. Our framework offers a flexible and powerful solution for real-world applications facing the dual challenge of covariate and semantic shifts, paving the way for more robust and generalizable machine learning models.

\begin{figure*}[!t]
    \centering
    \includegraphics[width=0.95\linewidth]{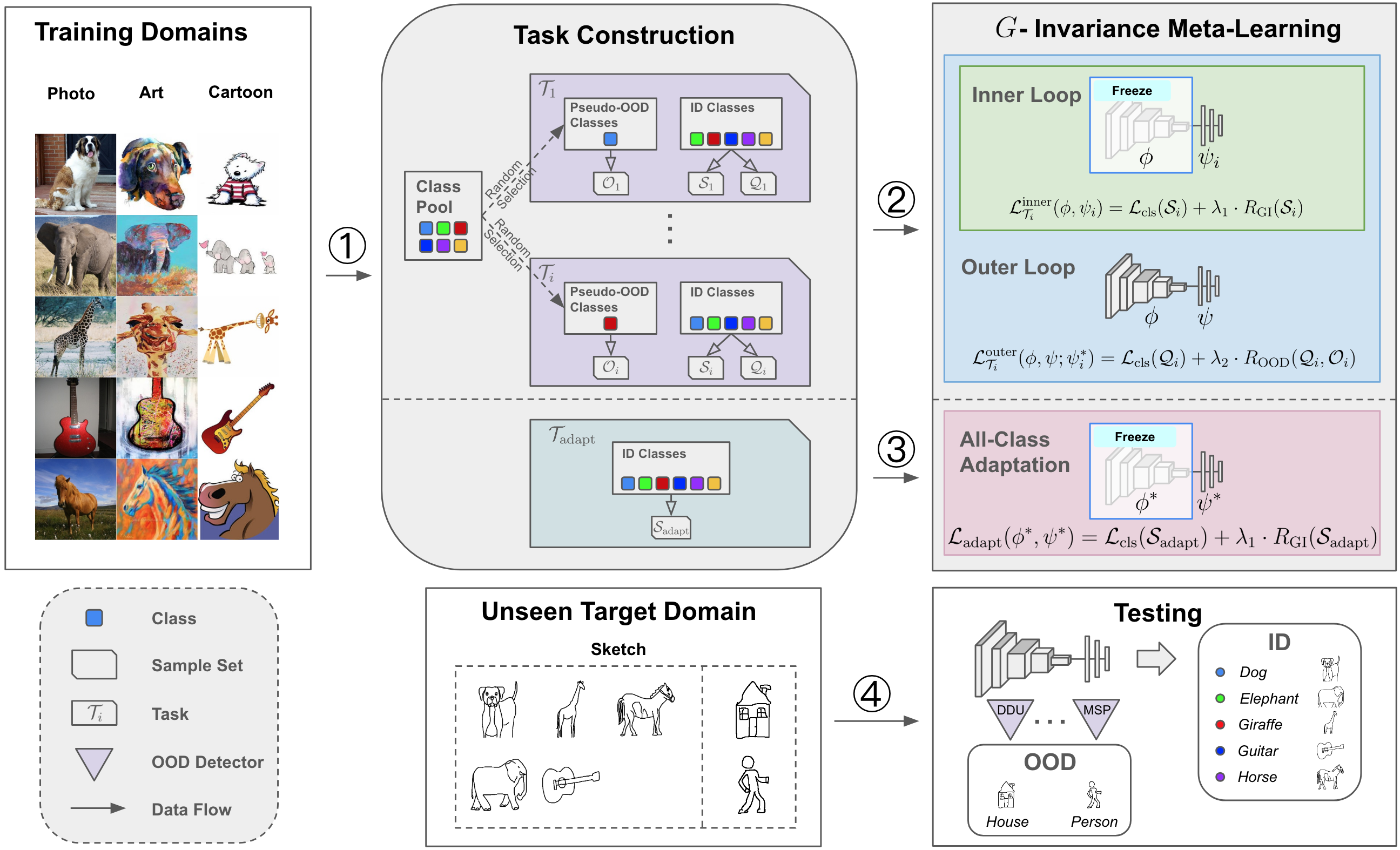}
    \caption{Overview of our proposed framework: Meta-learned Across Domain Out-of-distribution Detection (\sysname{}). The framework comprises four main stages: (1) Task construction: We randomly designate ID classes as pseudo-OOD classes for each task, simulating diverse OOD scenarios within the known class space. (2) Meta-learning: We perform bi-level optimization through inner and outer loop training. The inner loop focuses on task-specific adaptation across multiple domains, while the outer loop strengthens decision boundaries between ID and OOD, enhancing the model's sensitivity to distributional shifts. (3) All-class adaptation: We fine-tune the classification head on all ID classes while freezing the featurizer $\phi$, ensuring the classification head can effectively handle all in-distribution classes, including those previously used as pseudo-OODs, while preserving the learned domain-invariant features. (4) Test: We evaluate the model, trained on multiple source domains, on an unseen target domain for both ID classification and OOD detection. This framework aims to learn robust, domain-invariant features through meta-learning, achieving superior generalization to unseen target domains and effectively addressing the dual challenge of covariate and semantic shifts.}
    \label{fig:overview}
\end{figure*}

\section{Related Work}
    \textbf{Domain Generalization (DG):} Domain generalization techniques aim to capture semantic invariance across domains to model covariate shifts \cite{kamath2021does,wang2022toward,rosenfeld2020risks,li2018learning,ganin2016domain,balaji2018metareg}. Traditional methods like ERM \cite{vapnik1999nature} and IRM \cite{arjovsky2019invariant} focused on invariant feature learning but often struggled with significantly divergent test data. Recent strategies explore more sophisticated approaches, such as data augmentation methods like DDG \cite{zhang2022towards}, which aim to reduce semantic disparities while enhancing variation across domains. However, these approaches may not effectively separate semantics from variations due to limited training diversity. Another strategy involves creating new variations within augmented domains, like MBDG \cite{robey2021model}, which seeks domain-invariant feature extraction. Our work builds upon this concept, incorporating $G$-invariance within a meta-learning framework to better address both covariate and semantic shifts. Despite these advancements, conventional DG methods often neglect semantic OOD detection, and integrating traditional OOD techniques into DG frameworks frequently leads to suboptimal performance.

\textbf{OOD Detection:} Out-of-distribution detection encompasses tasks such as anomaly detection and open set recognition \cite{yang2021generalized}. Output-based methods, such as MSP \cite{hendrycks2016baseline}, ODIN \cite{liang2017enhancing}, and Energy \cite{liu2020energy}, manipulate the final layer's outputs, assuming OODs produce uniform class probabilities. Conversely, feature-based approaches, like the Mahalanobis distance score \cite{lee2018simple} and DDU \cite{mukhoti2023deep}, utilize intermediate layer features, offering richer information for OOD detection. Traditional OOD detection methods often fall short when confronted with covariate shifts, as they are typically trained on data from a single domain. This limitation leads to a fundamental flaw: these models are prone to incorrectly flagging samples from known classes as OOD merely due to shifts in the underlying data distribution, rather than true semantic differences.

Several methods approach OOD detection using meta-learning. OOD-MAML \cite{jeong2020ood} employs a GAN model to generate adversarial samples, using them as training OODs to train a K+1 classifier capable of directly classifying OOD instances. OEC \cite{wang2020few} adapts traditional OOD detection methods to a few-shot setting, aiming to detect out-of-episode and out-of-dataset OODs. Iwata et al. \cite{iwata2022meta} use meta-learning to train a neural network shared among all tasks, flexibly mapping instances from the original space to a latent space, then employing density estimation to detect OODs in this latent space. MOL \cite{wu2023meta} approaches this problem using both meta-learning and domain adaptation, learning a model on a sequence of ID distributions with different variations. It aims to quickly adapt to shifted ID distributions by fine-tuning a lightweight adapter and classification head using the trained initialization. However, except for MOL \cite{wu2023meta}, all these methods assume train and test data come from the same domain and are tested on a different set of classes not seen during training. In contrast, our problem involves test data from a different domain but with the same ID class labels. Additionally, in the test phase, all these methods require access to a few shots of ID test samples for adaptation, as is common in meta-learning methods. In our problem setting, however, the test domain is assumed to be unknown, and therefore, there is no access to any test data for model adaptation.

\textbf{Semantic OOD Detection in Unseen Domains:} Limited studies address semantic OOD detection in unseen domains. \cite{katsumata2021open} propose using triplet loss and decoupling loss to maximize the distance between different classes, including unknown ones, with the assumption that this separability will be maintained in unseen target domains. However, this approach requires OOD training samples of unknown labels, which are typically difficult to obtain. DAML \cite{shu2021open} approaches this problem by first augmenting each domain using cross-domain features and labels, then conducting meta-learning over the augmented domains to learn a generalizable open-domain solution. Building upon DAML's domain augmentation techniques, \cite{noguchi2023simple} argues that simple DG baselines like CORrelation ALignment (CORAL) and Maximum Mean Discrepancy (MMD) can perform comparably to DAML with reduced computation. However, both methods rely on specialized model architectures that are incompatible with state-of-the-art OOD detection methods. MEDIC \cite{wang2023generalizable} adopts meta-learning to develop a solution applicable across all domains but requires training multiple binary classifiers to reject unknown classes. SCONE \cite{bai2023feed} tackles this problem differently by using unlabeled wild data to directly train a binary OOD detector alongside the classification network. Both MEDIC and SCONE include a predefined OOD detector optimized with the classification backbone, limiting their adaptability to various OOD detection techniques. Our proposed \sysname{} framework addresses these limitations by leveraging meta-learning and $G$-invariance to capture domain-invariant features. This approach eliminates the need for specialized architectures while enabling seamless integration with various state-of-the-art OOD detection techniques.

\section{Preliminaries} 
    \subsection{Domain Generalization}
Domain generalization (DG) addresses the challenge of learning from multiple source domains to perform well on unseen target domains. Consider a feature space $\mathcal{X} \subseteq \mathbb{R}^d$ and a label space $\mathcal{Y} = \{1, \ldots, K\}$. The underlying data distribution $\mathbb{P}_{\text{ID}}(X, Y)$ is unknown and cannot be sampled directly. Instead, we observe data from a set of domains $\mathcal{E}_{\text{all}}$, where each domain $e \in \mathcal{E}_{\text{all}}$ has its own distribution $\mathbb{P}_{\text{ID}}(X^e, Y^e)$. 

Given training data from a subset of domains $\mathcal{E}_{\text{train}} \subset \mathcal{E}_{\text{all}}$, the objective is to learn a classifier $f_\theta: \mathcal{X} \to \mathcal{P}(\mathcal{Y})$ that performs well across all domains, including unseen ones. Here, $\mathcal{P}(\mathcal{Y})$ represents the space of probability distributions over $\mathcal{Y}$. This setting differs from traditional supervised learning as it requires generalization not just to unseen samples, but to entirely new domains with potentially different underlying distributions.

\subsection{Meta-Learning}
Conventional meta-learning aims to train a model that can quickly adapt to new tasks with limited data. Let $\mathcal{T} = {T_1, T_2, ..., T_N}$ be a set of tasks sampled from a task distribution $p(\mathcal{T})$. Each task $T_i$ consists of a support set $\mathcal{S}_i$ and a query set $\mathcal{Q}_i$. The meta-learning objective is to find parameters $\theta$ that minimize the expected loss across tasks:
\begin{equation}
\theta^* = \arg\min_{\theta} \mathbb{E}_{T_i \sim p(\mathcal{T})} \mathcal{L}(f_{\theta_i^*}; \mathcal{Q}_i)
\end{equation}
where $\theta_i^* = \theta - \alpha \nabla_{\theta} \mathcal{L}(f_{\theta}; \mathcal{S}_i)$ represents task-specific parameters adapted from $\theta$ using the support set $\mathcal{S}_i$, and $\alpha$ is the learning rate for the inner loop optimization. This formulation allows the model to learn a good initialization that can quickly adapt to new tasks, facilitating generalization across different domains or distributions. 

In contrast, our work adopts the meta-learning paradigm to expose the model to a wide variety of training distributions, aiming to improve its generalizability to unseen target domains and learn domain-invariant features that are effective for both ID classification and OOD detection.

\subsection{$G$-invariance and Domain Augmentation}
\label{sec:$G$-invariance}
We incorporate the concept of $G$-invariance into our meta-learning framework. Following the model-based domain generalization (MBDG) approach \cite{robey2021model}, we assume a domain transformation model $G: \mathcal{X} \times \mathcal{E}_{\text{all}} \to \mathcal{X}$ that relates the random variables $X$ and $X^e$ via $X\mapsto G(X,e) = X^e$. $G$-invariance ensures that the model $f_{\theta}$ satisfies:
\begin{equation}
f_{\theta}(\mathbf{x}) = f_{\theta}(G(\mathbf{x},e)) \quad \forall \mathbf{x} \in \mathcal{X}, e \in \mathcal{E}_{\text{all}}
\end{equation}

This property encourages the model to learn features that are invariant across domains, improving generalization to unseen domains.

The domain transformation model \(G\) is approximated through a Multimodal Image-to-Image Translation Network (MIITN) \cite{huang2018multimodal}, consisting of a disentangled encoder network \(E\) and a generative decoder network \(D\). The encoder \(E\) is composed of two parts: a semantic encoder \(E_s\) and a variation encoder \(E_v\). For a given data instance \(\mathbf{x}^e\) from domain \(e \in \mathcal{E}_{\text{all}}\), \(E_s\) extracts the conceptual vector \(\mathbf{s} = E_s(\mathbf{x}^e)\), while \(E_v\) extracts the variation vector \(\mathbf{v}^e = E_v(\mathbf{x}^e)\) encapsulating environmental factors such as lighting conditions or viewpoint.

To generate augmented domain images with random variations, the decoder \(D\) reconstructs a new data instance \(\mathbf{x}^{e'}\) using the conceptual vector \(\mathbf{s}\) and a potentially different style vector \(\mathbf{v}^{e'}\): \(\mathbf{x}^{e'} = D(\mathbf{s}, \mathbf{v}^{e'})\). Consequently, the domain transformation model \(G\) is formulated as:
\begin{equation}
G(\mathbf{x}, e) = D(E_s(\mathbf{x}), \mathbf{v}^e)
\end{equation}

\subsection{Semantic OOD Detection in Unseen Domains}
In addition to the ID probability distribution $\mathbb{P}_{\text{ID}}(X, Y)$, we consider an OOD probability distribution $\mathbb{P}_{\text{OOD}}(X_O, Y_O)$, where $X_O$ is a random variable from $\mathcal{X}$, but $Y_O$ is a random variable whose values belong to a set disjoint from $\mathcal{Y}$, i.e., $Y_O \in \mathcal{Y}_O$ where $\mathcal{Y}_O \cap \mathcal{Y} = \emptyset$. The goal is to learn a classifier $f_{\theta}$ that satisfies two objectives across all domains $\mathcal{E}_{\text{all}}$:

1) Accurately predict the labels $Y$ for instances $X$ drawn from $\mathbb{P}_{\text{ID}}(X, Y)$.
2) Enable an OOD detector $q_\omega({\bf x}; f_{\theta})$ to classify ID and OOD instances: $q_\omega({\bf x}_O) = \text{OOD}$ and $q_\omega({\bf x}_I) = \text{ID}$, for each $({\bf x}_I, y_I)\sim \mathbb{P}_{\text{ID}}(X, Y)$ and $({\bf x}_O, y_O) \sim \mathbb{P}_{\text{OOD}}(X_O, Y_O)$.

The OOD detector $q_\omega({\bf x}; f_{\theta}): \mathcal{X} \rightarrow \{\text{ID}, \text{OOD}\}$ is defined as:
\begin{equation}
q_\omega({\bf x}; f_{\theta}) = 
\begin{cases}
\text{OOD}, & \text{if } o({\bf x}; f_{\theta}) \ge \omega \\
\text{ID}, & \text{otherwise}
\end{cases}
\end{equation}

where $o({\bf x}; f_{\theta})$ is an OOD scoring function and $\omega$ is a given threshold.

\section{The \sysname{} Framework} 
    In this section, we present our Meta-learned Across Domain Out-of-distribution Detection (\sysname{}) framework. Our approach leverages meta-learning to enhance the model's ability to generalize across domains and detect OOD instances effectively. A key distinction of our framework from conventional meta-learning methods lies in the nature of our problem: we test on unseen target domains that share the same ID class set as the training domains. This characteristic allows us to construct tasks without using relative labels, unlike typical few-shot learning scenarios. Instead, we leverage this structure to create diverse tasks that simulate OOD scenarios using existing ID classes, enhancing the model's generalization capabilities.

\subsection{Overview of \sysname{}}

\sysname{} consists of three main components:
\begin{enumerate}
    \item A meta-learning framework for task construction and model training;
    \item $G$-invariance regularization for domain generalization;
    \item An OOD regularization term for enhancing semantic OOD detection;
\end{enumerate}

The overall training procedure is outlined in Algorithm \ref{alg:our_alg}.

\begin{algorithm}[!t]
\small
    \caption{\sysname{} Training Procedure}
    \label{alg:our_alg}
    \begin{algorithmic}[1]
    \While{not done}
        \State Sample tasks $\{\mathcal{T}_i\}_{i=1}^L$ with random pseudo-OOD classes
        \For{each task $\mathcal{T}_i$}
            % \State Create support set $\mathcal{B}_{\text{support}}$ and augmented support set $\mathcal{B}_{\text{support}}^{\text{aug}}$
            \State Calculate classification loss $\mathcal{L}_{cls}$ on the support set $\mathcal{S}_i$
            \State Create augmented inputs $\mathcal{S}_i^{\text{aug}}=\{(\Tilde{\mathbf{x}}_i,y_{\text{in}})\}_{i=1}^M$ using random styles: $(\Tilde{\mathbf{x}}_i,y_{\text{in}})\leftarrow\textsc{DataAug}(\mathbf{x}_{\text{in}},y_{\text{in}}),\: \forall i\in[M]$
            \State Calculate $G$-invariance regularization $R_{\text{GI}}$ using $\mathcal{S}_i$ and $\mathcal{S}_i^{\text{aug}}$ (Eq. \ref{eq:gi-reg})
            \State Compute task-specific loss $\mathcal{L}_{\mathcal{T}_i}^{\text{inner}}$ (Eq. \ref{eq:inner-loss})
            \State Update task-specific parameters $\psi_i^{'}$ (Eq. \ref{eq:inner-update})
            % \State Create ID query set $\mathcal{B}_{\text{query}}^{\text{ID}}$ and OOD query set $\mathcal{B}_{\text{query}}^{\text{OOD}}$
            \State Calculate OOD regularization $R_{\text{OOD}}$ using $\mathcal{Q}_i$ and $\mathcal{O}_i$ (Eq. \ref{eq:ood-reg})
            \State Compute outer loss $\mathcal{L}_{\mathcal{T}_i}^{\text{outer}}$ (Eq. \ref{eq:outer-loss})
        \EndFor
        \State Update global parameters $(\phi,\psi)$ (Eq. \ref{eq:outer-update})
    \EndWhile
    \State Perform all-class adaptation of classification head $h$ (Eq. \ref{eq:adapt-loss} and Eq. \ref{eq:adapt-update})

    \State            

    \Procedure{\textsc{DataAug}}{$\mathbf{x},y$}
        \State $\mathbf{s}\leftarrow E_s(\mathbf{x})$, $\mathbf{v}\leftarrow E_v(\mathbf{x})$
        \State Sample $\mathbf{v}'\sim\mathcal{N}(0,\mathbf{I})$
        \State \textbf{return } $(D(\mathbf{s},\mathbf{v}'),y)$
    \EndProcedure
    \end{algorithmic}
\end{algorithm}

\subsection{Meta-learning for Domain Generalization and OOD Detection}

In this work, we leverage meta-learning to expose our model to a diverse set of training distributions/tasks. Our goal is to develop a classifier $f_{\theta}: \mathcal{X} \to \mathcal{P}(\mathcal{Y})$ that can effectively generalize to unknown test domains $\mathcal{E}_{\text{all}} \setminus \mathcal{E}_{\text{train}}$, accurately classifying ID samples while producing features suitable for downstream OOD detection. The classifier $f_{\theta}$ is decomposed into a featurizer $g_{\phi}: \mathcal{X} \to \mathcal{Z}$ parameterized by $\phi$, and a classification head $h_{\psi}: \mathcal{Z} \to \mathcal{P}(\mathcal{Y})$ parameterized by $\psi$, such that $f_{\theta}(\mathbf{x}) = h_{\psi}(g_{\phi}(\mathbf{x}))$. By training on a variety of tasks, we aim to learn a model whose intermediate representations (outputs of $g_{\phi}$) and final outputs (from $h_{\psi}$) are both discriminative for ID classification and adaptable for effective OOD detection in unseen target domains.

We split the dataset $\mathcal{D}$ into two sets:
\begin{itemize}
\item $\mathcal{D}_{\text{train}}$: containing all the training domains $e \in \mathcal{E}_{\text{train}}$
\item $\mathcal{D}_{\text{test}}$: containing all the test domains $e \in \mathcal{E}_{\text{all}} \setminus \mathcal{E}_{\text{train}}$
\end{itemize}

We sample a batch of $L$ tasks $\{\mathcal{T}_i\}_{i=1}^L$ from a task distribution $p(\mathcal{T})$. Each task $\mathcal{T}_i$ is defined as:
\begin{equation}
\mathcal{T}_i = (\mathcal{S}_i, \mathcal{Q}_i, \mathcal{O}_i)
\end{equation}
where:
\begin{itemize}
\item $\mathcal{S}_i \subset \mathcal{D}_{\text{train}}$ is the support set used for task adaptation
\item $\mathcal{Q}_i \subset \mathcal{D}_{\text{train}}$ is the query set used for meta-update
\item $\mathcal{O}_i \subset \mathcal{D}_{\text{train}}$ is the pseudo-OOD set used for meta-update
\end{itemize}

Importantly, $\mathcal{S}_i$, $\mathcal{Q}_i$, and $\mathcal{O}_i$ are randomly sampled across all training domains in $\mathcal{E}_{\text{train}}$, ensuring diversity in domain representation for each task. For each task $\mathcal{T}_i$, we randomly select a subset of classes from $\mathcal{Y}$ to serve as pseudo-OOD classes. The samples belonging to these selected classes form $\mathcal{O}_i$, while $\mathcal{S}_i$ and $\mathcal{Q}_i$ contain samples from the remaining classes. This ensures that:

\begin{itemize}
\item $\mathcal{S}_i \cap \mathcal{Q}_i = \emptyset$, $\mathcal{S}_i \cap \mathcal{O}_i = \emptyset$, $\mathcal{Q}_i \cap \mathcal{O}_i = \emptyset$
\item The classes in $\mathcal{S}_i$ and $\mathcal{Q}_i$ have no overlap with the classes in $\mathcal{O}_i$
\item Different tasks $\mathcal{T}_i$ have different randomly selected pseudo-OOD classes
\end{itemize}

This task construction method offers several interconnected benefits. Primarily, it significantly enhances task diversity, exposing the model to a wide variety of OOD scenarios without the need for external OOD data. This approach leverages existing ID data, eliminating the often challenging and impractical process of collecting or generating external OOD samples. By treating different ID classes as OODs during training, the model is compelled to learn more general characteristics for OOD detection rather than relying on specific class features. This not only promotes better generalization to truly unseen OOD samples but also enhances the model's ability to make fine-grained distinctions. The model is forced to learn highly discriminative features that can differentiate between closely related classes (ID vs. pseudo-OOD), potentially leading to more robust and nuanced feature representations overall. This comprehensive approach to task construction and feature learning lays a strong foundation for improved OOD detection and generalization capabilities.

\subsection{$G$-invariance Regularization}

To enhance domain generalization, we incorporate $G$-invariance regularization into our meta-learning framework. The $G$-invariance term $R_{\text{GI}}$ is defined as:
\begin{equation}
R_{\text{GI}}(\phi, \psi_i; \mathcal{S}_i) = \frac{1}{|\mathcal{S}_i|}\sum_{(\mathbf{x}_{\text{in}}, y_{\text{in}}) \in \mathcal{S}_i} d[h_{\psi_i}(g_\phi(\mathbf{x}_{\text{in}})), h_{\psi_i}(g_\phi(\Tilde{\mathbf{x}}_{\text{in}}))]
\label{eq:gi-reg}
\end{equation}

Here, $d[\cdot,\cdot]$ denotes a distance metric; in this work, we use the $\ell_1$ norm. $g_\phi$ represents the featurizer, and $h_{\psi_i}$ is the task-specific classification head. The term $\Tilde{\mathbf{x}}_i$ refers to an augmented version of $\mathbf{x}_{\text{in}}$ with random domain variation, as detailed in Section~\ref{sec:$G$-invariance}. This regularization term encourages the model to learn domain-invariant features, thereby improving generalization across different domains.

\subsection{OOD Regularization}

To improve semantic OOD detection, we introduce an OOD regularization term $R_{\text{OOD}}$ based on an energy-based scoring approach \cite{liu2020energy}:
\begin{equation}
\begin{aligned}
R_{\text{OOD}}(\phi, \psi_i; \mathcal{Q}_i, \mathcal{O}_i) = &\sum_{(\mathbf{x}_{\text{in}}, y_{\text{in}}) \in \mathcal{Q}_i, (\mathbf{x}_{\text{out}}, y_{\text{out}}) \in \mathcal{O}_i} \\
&\big[ \max(0,E_g(\mathbf{x}_{\text{in}},\phi,\psi_i)-m_{I}) \big]^2 + \\ 
&\big[ \max(0,m_{O}-E_g(\mathbf{x}_{\text{out}},\phi,\psi_i)) \big]^2
\end{aligned}
\label{eq:ood-reg}
\end{equation}

where $m_{I}$ and $m_{O}$ are margins that regularize the energy scores for ID and OOD instances, respectively. The energy function $E_g: \mathcal{X} \times \Theta \to \mathbb{R}$ is defined as:
\begin{equation}
E_g(\mathbf{x}, \phi, \psi) = -T \cdot \log \sum_{k=1}^K \exp\left(\frac{\text{logit}_k (f_{\phi, \psi}(\mathbf{x}))}{T}\right)
\end{equation}
with $T$ denoting the temperature parameter and $\text{logit}_k (f_{\phi, \psi}(\mathbf{x}))$ representing the logit corresponding to the $k$-th class label.

\subsection{\sysname{} Model Learning}

The learning problem of our proposed \sysname{} framework is formulated as a nested optimization problem inspired by bi-level optimization \cite{liu2021investigating} and meta-learning \cite{hospedales2021meta} principles:
\begin{equation}
(\phi^*, \psi^*) = \arg\min_{(\phi, \psi)} \mathbb{E}_{\mathcal{T}_i \sim p(\mathcal{T})} \mathcal{L}_{\mathcal{T}_i}^{\text{outer}}(\phi, \psi; \psi_i^*, \mathcal{Q}_i, \mathcal{O}_i)
\end{equation}

where $\psi_i^*$ is the solution to the inner optimization problem:
\begin{equation}
\psi_i^* = \arg\min_{\psi_i} \mathcal{L}_{\mathcal{T}_i}^{\text{inner}}(\phi, \psi_i; \mathcal{S}_i)
\end{equation}

The inner optimization problem aims to learn task-specific parameters $\psi_i^*$ that minimize the inner loss $\mathcal{L}_{\mathcal{T}_i}^{\text{inner}}$ for each task $\mathcal{T}_i$ using the support set $\mathcal{S}_i$. This process adapts the model to specific tasks within the meta-learning framework. While the outer optimization problem seeks to find global parameters $(\phi^, \psi^)$ that minimize the expected outer loss $\mathcal{L}_{\mathcal{T}_i}^{\text{outer}}$ across all tasks. This loss includes both classification performance on the query set $\mathcal{Q}_i$ and an OOD regularization term. The outer problem aims to learn parameters that generalize well across tasks and domains, improving both in-distribution classification and OOD detection.

\textbf{Inner Loss:} The inner loss is defined as:
\begin{equation}
\mathcal{L}_{\mathcal{T}_i}^{\text{inner}}(\phi, \psi_i; \mathcal{S}_i) = \mathcal{L}_{\text{cls}}(\phi, \psi_i; \mathcal{S}_i) + \lambda_1 \cdot R_{\text{GI}}(\phi, \psi_i; \mathcal{S}_i)
\label{eq:inner-loss}
\end{equation}

where $\mathcal{L}_{\text{cls}}$ is the classification loss and $R_{\text{GI}}$ is the $G$-invariance regularization term.

\textbf{Outer Loss:} The outer loss is defined as:
\begin{equation}
\begin{split}
\mathcal{L}_{\mathcal{T}_i}^{\text{outer}}(\phi, \psi; \psi_i^*, \mathcal{Q}_i, \mathcal{O}_i) = \mathcal{L}_{\text{cls}}(\phi, \psi_i^*; \mathcal{Q}_i) \\
    + \lambda_2 \cdot R_{\text{OOD}}(\phi, \psi_i^*; \mathcal{Q}_i, \mathcal{O}_i)
\end{split}
\label{eq:outer-loss}
\end{equation}

where $\mathcal{L}_{\text{cls}}$ is the classification loss on the query set, and $R_{\text{OOD}}$ is the OOD regularization term based on energy scores.

In practice, we solve these optimization problems using stochastic gradient descent (SGD) as detailed in Algorithm \ref{alg:our_alg}. The inner loop optimization updates the task-specific parameters $\psi_i$:
\begin{equation}
\psi_i^{\prime} = \psi_i - \eta_1 \cdot \nabla_{\psi_i} \mathcal{L}_{\mathcal{T}_i}^{\text{inner}}(\phi, \psi_i; \mathcal{S}_i)
\label{eq:inner-update}
\end{equation}

The outer loop optimization updates the global parameters $(\phi, \psi)$:
\begin{equation}
(\phi, \psi) = (\phi, \psi) - \eta_2 \cdot \nabla_{(\phi, \psi)} \sum_{\mathcal{T}_i \sim p(\mathcal{T})} \mathcal{L}_{\mathcal{T}_i}^{\text{outer}}(\phi, \psi; \psi_i^{\prime}, \mathcal{Q}_i, \mathcal{O}_i)
\label{eq:outer-update}
\end{equation}

After the main training loop, we perform an all-class adaptation of the classification head to ensure it can handle all classes effectively, including those previously used as pseudo-OODs during training.

Note that this problem formulation differs from common meta-learning problems in two crucial aspects. First, we assume no access to any test domain data during training or inference. Consequently, there is no adaptation phase to the test data, which is typically present in traditional meta-learning setups. Second, unlike conventional few-shot learning scenarios, our tasks do not use relative labels. Instead, we leverage the fact that our test domains share the same ID class set as the training domains to construct tasks that simulate OOD scenarios using existing ID classes. This approach allows \sysname{} to learn a robust, domain-invariant model that can generalize directly to unseen domains without further adaptation, making it particularly suitable for real-world scenarios where test domain data is unavailable or inaccessible during the learning process.

\subsection{All-class Adaptation}

After obtaining the optimal parameters $(\phi^*, \psi^*)$ from the main training loop, we perform an all-class adaptation of the classification head. This additional step ensures the classification head can effectively handle all in-distribution classes, including those previously used as pseudo-OODs during training.

We define an adaptation task $\mathcal{T}_{\text{adapt}} = (\mathcal{S}_{\text{adapt}})$, where $\mathcal{S}_{\text{adapt}} \subset \mathcal{D}_{\text{train}}$ is the support set for adaptation, sampled from all classes in $\mathcal{D}_{\text{train}}$ without any pseudo-OOD holdouts.

Similar to the inner loop procedure, We freeze the featurizer $g_{\phi^*}$ and update only the classification head, starting from $h_{\psi^*}$. The adaptation loss is defined as:
\begin{equation}
\begin{split}
\mathcal{L}_{\text{adapt}}(\phi^*, \psi^*; \mathcal{S}_{\text{adapt}}) = \mathcal{L}_{\text{cls}}(\phi^*, \psi^*; \mathcal{S}_{\text{adapt}}) \\
+ \lambda_1 \cdot R_{\text{GI}}(\phi^*, \psi^*; \mathcal{S}_{\text{adapt}})
\end{split}
\label{eq:adapt-loss}
\end{equation}

where $\mathcal{L}_{\text{cls}}$ is the classification loss and $R_{\text{GI}}$ is the $G$-invariance regularization term, calculated similarly to the inner loop but using all ID classes.

We update the classification head parameters using gradient descent:
\begin{equation}
\psi^{\text{final}} = \psi^* - \eta_1 \cdot \nabla_{\psi} \mathcal{L}_{\text{adapt}}(\phi^*, \psi^*; \mathcal{S}_{\text{adapt}})
\label{eq:adapt-update}
\end{equation}

where $\psi^{\text{final}}$ represents the final adapted classification head parameters after this additional step.

This adaptation process ensures that the classification head maintains its performance on all in-distribution classes while preserving the domain-invariant features learned during the main training phase.

\section{Experiments}
    \subsection{Experimental Setting}

\textbf{Datasets.} We evaluate \sysname{} on four popular datasets widely used in domain generalization research: \textsc{ColoredMNIST} \cite{arjovsky2019invariant} (3 domains, 70,000 samples, 2 classes), \textsc{PACS} \cite{li2017deeper} (4 domains, 9,991 samples, 7 classes), \textsc{VLCS} \cite{torralba2011unbiased} (4 domains, 10,729 samples, 5 classes), and \textsc{TerraIncognita} \cite{beery2018recognition} (4 domains, 16,110 samples, 10 classes). Our experiments are conducted using the DomainBed framework \cite{gulrajani2020search} to ensure consistent assessment.

\begin{table}[!t]
\centering
\caption{Dataset Statistics.}
\label{tab:dataset-stats}
\begin{tabular}{lccc}
\hline
Dataset & \# Domains & \# Samples & \# Classes \\
\hline
\textsc{ColoredMNIST} & 3 & 70,000 & 2 \\
\textsc{PACS} & 4 & 9,991 & 7 \\
\textsc{VLCS} & 4 & 10,729 & 5 \\
\textsc{TerraIncognita} & 4 & 16,110 & 10 \\
\hline
\end{tabular}
\label{tab:dataset-statistic}
\end{table}

\textbf{Experimental Setup.}
The environment used: Python 3.9.13, PyTorch 1.12.1, Torchvision 0.13.1, CUDA 11.6, CUDNN 8302, NumPy 1.23.1, and PIL 9.2.0. Our experimental setup is tailored for semantic OOD detection in unseen domains. Crucially, OOD samples are always sourced from an unseen test domain. For each dataset, we designate one class as OOD and iterate our experiments with different randomly chosen OOD classes until a minimum of 40\% of classes have been designated as OOD. To mitigate the impact of randomness, each set of experiments is conducted in two separate trials with different seeds. The results reported represent mean and standard error across these trials and all OOD class choices.

During the training phase, one random class is deliberately held out as a pseudo-OOD class in each task. However, during model evaluation, all classes except for the true OOD class are considered ID. For conciseness, we report the averaged OOD detection performance across all trials and OOD class selections unless specified otherwise.

\textbf{Backbone Model.} The classifier $f$ in \sysname{} consists of two components: a featurizer $g$ and a classification head $h$. We implement $g$ as a ResNet50 architecture \cite{he2016deep} pre-trained on ImageNet, while $h$ is a fully connected layer with an input size of 2048 and an output size of $K$, where $K$ represents the number of classes. During training, we load a pre-trained featurizer $g$ and initialize the classification head $h$ with random weights. The input resolution is 224 by 224 for all datasets except for \textsc{ColoredMNIST}, which uses 32 by 32.

\textbf{OOD Detectors and Baselines.} We evaluate \sysname{} in combination with three commonly used OOD detection algorithms: DDU \cite{mukhoti2023deep}, MSP \cite{hendrycks2016baseline}, and Energy \cite{liu2020energy}. These methods exploit different features and heuristics for OOD detection. We compare \sysname{} against four recognized domain generalization baselines: ERM \cite{vapnik1999nature}, IRM \cite{arjovsky2019invariant}, Mixup \cite{yan2020improve}, and MBDG \cite{robey2021model}. Additionally, we include five SOTA open-set domain generalization baselines: DAML \cite{shu2021open}, MEDIC \cite{wang2023generalizable}, SCONE \cite{bai2023feed}, EDir-MMD and EDst-MMD \cite{noguchi2023simple}, to provide a comprehensive evaluation of \sysname{} in terms of both OOD detection and ID classification accuracy.

\textbf{Model Selection.} To balance experimental rigor with practical execution time constraints, we employ a leave-one-domain-out cross-validation strategy. This approach more accurately reflects real-world scenarios by assuming that training and test domains follow a meta-distribution over domains. For each algorithm and test domain, we conduct a random search of 20 trials across the hyperparameter distribution. Our hyperparameter optimization focuses primarily on $m_I$, $m_O$, $\eta_1$, and $\eta_2$. We typically use uniform distributions: Unif(-11.0, -9.0) for $m_I$, Unif(-9.0, -7.0) for $m_O$, Unif(-4.5, -2.5) for $\eta_1$, and Unif(-5.5, -4.0) for $\eta_2$. We set $T=1$ and $\lambda_1=\lambda_2=0.1$ across all experiments. For each meta-batch, we sample 4 tasks, with each task following an N-way-5-shot sampling scheme. We tailor specific ranges and default values to each dataset's unique characteristics. For a comprehensive breakdown of hyperparameter settings, please refer to our GitHub repository \href{https://github.com/haoliangwang86/MADOD}{[link]}.

\textbf{Evaluation Metrics.} We evaluate \sysname{}'s performance using standard metrics for OOD detection and classification accuracy. For OOD detection, we report the Area Under the Receiver Operating Characteristic curve (AUROC) and the Area Under the Precision-Recall Curve (AUPR). For ID classification accuracy, we report the standard classification accuracy on the ID test set.

\begin{table*}[!t]
\centering
\setlength\tabcolsep{4pt}
% \scriptsize
\caption{Performance evaluation. The top-performing scores are highlighted in \textbf{bold}, and the second-best scores are \underline{underlined}.}
\label{tab:summary-all}
\resizebox{\textwidth}{!}{%
\begin{tabular}{l|ccc|ccc|ccc|ccc}
\toprule
\multirow{2}{*}{Methods} & \multicolumn{3}{c|}{\textsc{ColoredMNIST}} & \multicolumn{3}{c|}{\textsc{PACS}} & \multicolumn{3}{c|}{\textsc{VLCS}} & \multicolumn{3}{c}{\textsc{TerraIncognita}} \\
\cmidrule{2-13}
 & \textbf{AUROC} & \textbf{AUPR} & \textbf{Accuracy} & \textbf{AUROC} & \textbf{AUPR} & \textbf{Accuracy} & \textbf{AUROC} & \textbf{AUPR} & \textbf{Accuracy} & \textbf{AUROC} & \textbf{AUPR} & \textbf{Accuracy} \\
\midrule
ERM* \cite{vapnik1999nature} & 52.17\phantom{0} $\pm$ \phantom{0}2.6 & 10.72\phantom{0} $\pm$ \phantom{0}0.4 & 34.71\phantom{0} $\pm$ 12.3 & 80.12\phantom{0} $\pm$ \phantom{0}1.9 & 38.96\phantom{0} $\pm$ \phantom{0}3.4 & 84.48\phantom{0} $\pm$ \phantom{0}4.8 & 72.04\phantom{0} $\pm$ \phantom{0}7.9 & 27.63\phantom{0} $\pm$ 12.3 & \textbf{80.14\phantom{0}} $\pm$ \phantom{0}7.2 & 50.56\phantom{0} $\pm$ \phantom{0}1.1 & 16.24\phantom{0} $\pm$ \phantom{0}4.1 & 52.2\phantom{0} $\pm$ \phantom{0}2.9 \\
IRM* \cite{arjovsky2019invariant} & 52.99\phantom{0} $\pm$ \phantom{0}0.2 & 11.15\phantom{0} $\pm$ \phantom{0}0.1 & 41.52\phantom{0} $\pm$ 14.5 & 81.07\phantom{0} $\pm$ \phantom{0}4.5 & \underline{42.14\phantom{0}} $\pm$ \phantom{0}7.6 & 83.92\phantom{0} $\pm$ \phantom{0}4.5 & 72.54\phantom{0} $\pm$ \phantom{0}6.5 & 27.82\phantom{0} $\pm$ 12.3 & 78.53\phantom{0} $\pm$ \phantom{0}7.2 & 54.18\phantom{0} $\pm$ \phantom{0}1.1 & 17.87\phantom{0} $\pm$ \phantom{0}5.5 & 52.0\phantom{0} $\pm$ \phantom{0}3.5 \\
Mixup* \cite{yan2020improve} & 53.23\phantom{0} $\pm$ \phantom{0}1.4 & 10.94\phantom{0} $\pm$ \phantom{0}0.8 & 36.44\phantom{0} $\pm$ 13.1 & 74.70\phantom{0} $\pm$ \phantom{0}4.5 & 35.22\phantom{0} $\pm$ \phantom{0}7.1 & \underline{84.92\phantom{0}} $\pm$ \phantom{0}5.0 & 68.18\phantom{0} $\pm$ \phantom{0}7.6 & 26.63\phantom{0} $\pm$ \phantom{0}4.6 & \underline{79.94\phantom{0}} $\pm$ \phantom{0}6.9 & 55.00\phantom{0} $\pm$ \phantom{0}1.7 & 16.84\phantom{0} $\pm$ \phantom{0}5.7 & \textbf{54.4\phantom{0}} $\pm$ \phantom{0}3.6 \\
MBDG* \cite{robey2021model} & 62.90\phantom{0} $\pm$ \phantom{0}7.7 & 15.16\phantom{0} $\pm$ \phantom{0}2.6 & 52.87\phantom{0} $\pm$ 16.8 & 78.51\phantom{0} $\pm$ \phantom{0}3.7 & 36.34\phantom{0} $\pm$ \phantom{0}5.0 & 79.40\phantom{0} $\pm$ \phantom{0}2.2 & 71.89\phantom{0} $\pm$ \phantom{0}5.9 & 21.16\phantom{0} $\pm$ \phantom{0}7.4 & 77.84\phantom{0} $\pm$ \phantom{0}7.6 & 52.78\phantom{0} $\pm$ \phantom{0}1.4 & 16.81\phantom{0} $\pm$ \phantom{0}4.4 & 44.0\phantom{0} $\pm$ \phantom{0}3.7 \\

EDir-MMD \cite{noguchi2023simple} & 47.30\phantom{0} $\pm$ \phantom{0}3.1 & 32.60\phantom{0} $\pm$ 22.7 & 60.62\phantom{0} $\pm$ \phantom{0}1.7 & 77.36\phantom{0} $\pm$ \phantom{0}1.7 & 32.85\phantom{0} $\pm$ \phantom{0}1.6 & 80.05\phantom{0} $\pm$ \phantom{0}0.3 & 55.66\phantom{0} $\pm$ \phantom{0}8.9 & 12.18\phantom{0} $\pm$ \phantom{0}0.6 & 72.82\phantom{0} $\pm$ \phantom{0}0.1 & 54.39\phantom{0} $\pm$ 14.8 & 18.50\phantom{0} $\pm$ \phantom{0}3.0 & 53.33\phantom{0} $\pm$ \phantom{0}1.8 \\
EDst-MMD \cite{noguchi2023simple} & 52.42\phantom{0} $\pm$ \phantom{0}0.6 & 34.64\phantom{0} $\pm$ 23.9 & \underline{60.97\phantom{0}} $\pm$ \phantom{0}2.8 & 77.80\phantom{0} $\pm$ \phantom{0}1.3 & 32.83\phantom{0} $\pm$ \phantom{0}2.1 & 81.44\phantom{0} $\pm$ \phantom{0}0.2 & 53.16\phantom{0} $\pm$ \phantom{0}5.9 & 11.01\phantom{0} $\pm$ \phantom{0}2.4 & 74.06\phantom{0} $\pm$ \phantom{0}1.2 & 54.06\phantom{0} $\pm$ 14.4 & 18.51\phantom{0} $\pm$ \phantom{0}3.7 & \underline{53.99\phantom{0}} $\pm$ \phantom{0}2.5 \\
SCONE \cite{bai2023feed} & 58.67\phantom{0} $\pm$ \phantom{0}4.0 & 22.13\phantom{0} $\pm$ \phantom{0}1.3 & 54.89\phantom{0} $\pm$ 16.1 & 71.31\phantom{0} $\pm$ \phantom{0}3.5 & 30.64\phantom{0} $\pm$ \phantom{0}8.5 & 82.06\phantom{0} $\pm$ \phantom{0}0.5 & 64.87\phantom{0} $\pm$ \phantom{0}1.2 & 24.40\phantom{0} $\pm$ \phantom{0}3.5 & 77.96\phantom{0} $\pm$ \phantom{0}2.2 & 51.13\phantom{0} $\pm$ \phantom{0}8.7 & 16.30\phantom{0} $\pm$ \phantom{0}4.2 & 50.96\phantom{0} $\pm$ \phantom{0}3.2 \\
DAML \cite{shu2021open} & 51.82\phantom{0} $\pm$ \phantom{0}0.4 & 34.85\phantom{0} $\pm$ \phantom{0}0.1 & 53.43\phantom{0} $\pm$ 21.6 & 73.42\phantom{0} $\pm$ \phantom{0}0.8 & 31.29\phantom{0} $\pm$ \phantom{0}8.2 & 80.82\phantom{0} $\pm$ \phantom{0}0.4 & 65.91\phantom{0} $\pm$ \phantom{0}0.0 & 22.84\phantom{0} $\pm$ \phantom{0}2.2 & 78.45\phantom{0} $\pm$ \phantom{0}0.6 & 55.86\phantom{0} $\pm$ \phantom{0}9.9 & 16.17\phantom{0} $\pm$ \phantom{0}3.7 & 47.43\phantom{0} $\pm$ \phantom{0}2.6 \\
MEDIC \cite{wang2023generalizable} & 60.95\phantom{0} $\pm$ \phantom{0}4.6 & 36.31\phantom{0} $\pm$ \phantom{0}3.1 & 60.36\phantom{0} $\pm$ 15.4 & 76.95\phantom{0} $\pm$ \phantom{0}4.1 & 35.85\phantom{0} $\pm$ 10.5 & 83.77\phantom{0} $\pm$ \phantom{0}0.3 & 72.83\phantom{0} $\pm$ \phantom{0}0.5 & \underline{31.38\phantom{0}} $\pm$ \phantom{0}1.4 & 78.24\phantom{0} $\pm$ \phantom{0}0.9 & 56.67\phantom{0} $\pm$ \phantom{0}6.5 & 16.95\phantom{0} $\pm$ \phantom{0}4.8 & 46.80\phantom{0} $\pm$ \phantom{0}1.0 \\
\cmidrule{1-13}
\sysname{}-DDU & 60.71\phantom{0} $\pm$ \phantom{0}5.7 & 37.03\phantom{0} $\pm$ \phantom{0}0.8 & \textbf{68.60\phantom{0}} $\pm$ \phantom{0}3.5 & \textbf{88.68\phantom{0}} $\pm$ \phantom{0}3.3 & \textbf{50.91\phantom{0}} $\pm$ 13.4 & \textbf{85.38\phantom{0}} $\pm$ \phantom{0}3.9 & \textbf{79.71\phantom{0}} $\pm$ \phantom{0}6.2 & \textbf{36.35\phantom{0}} $\pm$ \phantom{0}9.1 & 78.53\phantom{0} $\pm$ \phantom{0}6.8 & \textbf{60.70\phantom{0}} $\pm$ \phantom{0}2.0 & \underline{20.28\phantom{0}} $\pm$ \phantom{0}3.5 & 53.06\phantom{0} $\pm$ \phantom{0}4.4 \\
\sysname{}-MSP & \underline{65.03\phantom{0}} $\pm$ \phantom{0}3.9 & \underline{39.17\phantom{0}} $\pm$ \phantom{0}2.6 & \textbf{68.60\phantom{0}} $\pm$ \phantom{0}3.5 & 81.20\phantom{0} $\pm$ \phantom{0}3.6 & 34.12\phantom{0} $\pm$ \phantom{0}8.2 & \textbf{85.38\phantom{0}} $\pm$ \phantom{0}3.9 & \underline{76.03\phantom{0}} $\pm$ \phantom{0}5.3 & 28.52\phantom{0} $\pm$ \phantom{0}9.6 & 78.53\phantom{0} $\pm$ \phantom{0}6.8 & \underline{60.39\phantom{0}} $\pm$ \phantom{0}2.7 & 18.42\phantom{0} $\pm$ \phantom{0}5.1 & 53.06\phantom{0} $\pm$ \phantom{0}4.4 \\
\sysname{}-Energy & \textbf{65.17\phantom{0}} $\pm$ \phantom{0}4.1 & \textbf{39.39\phantom{0}} $\pm$ \phantom{0}2.8 & \textbf{68.60\phantom{0}} $\pm$ \phantom{0}3.5 & \underline{81.78\phantom{0}} $\pm$ \phantom{0}3.6 & 37.44\phantom{0} $\pm$ \phantom{0}8.0 & \textbf{85.38\phantom{0}} $\pm$ \phantom{0}3.9 & 74.36\phantom{0} $\pm$ \phantom{0}7.0 & 29.40\phantom{0} $\pm$ 10.1 & 78.53\phantom{0} $\pm$ \phantom{0}6.8 & 57.83\phantom{0} $\pm$ \phantom{0}1.8 & \textbf{20.54\phantom{0}} $\pm$ \phantom{0}3.5 & 53.06\phantom{0} $\pm$ \phantom{0}4.4 \\
\bottomrule
\end{tabular}%
}
\\[3pt] % Adds vertical space after the table
\raggedright % Aligns text to the left
\textit{* The reported values for these baselines represent their best performance across all three OOD detectors: DDU, MSP, and Energy.}
\end{table*}

\subsection{Results}

\textbf{Performance Evaluation.} Table \ref{tab:summary-all} presents a comprehensive comparison of OOD detection performance between \sysname{} and various domain generalization and open-set domain generalization baselines.

\sysname{} demonstrates superior performance across different OOD detection algorithms and datasets, significantly outperforming both traditional domain generalization methods and state-of-the-art open-set domain generalization approaches. In terms of AUROC scores, \sysname{} consistently achieves top performance across all datasets. For example, on \textsc{ColoredMNIST}, \sysname{}-Energy reaches an AUROC of 65.17\%, surpassing the best baseline (MBDG) by 2.27\%. On \textsc{PACS}, \sysname{}-DDU records an AUROC of 88.68\%, exceeding the best baseline (IRM) by 7.61\%. Similar improvements are seen on \textsc{VLCS} and \textsc{TerraIncognita}, where \sysname{}-DDU outperforms the best baselines by 6.88\% and 4.03\%, respectively.

The gains in AUPR are even more significant. On \textsc{PACS}, \sysname{}-DDU achieves an AUPR of 50.91\%, beating the best baseline (IRM) by 8.77\%. On \textsc{VLCS}, it leads with 36.35\%, outperforming the top baseline (MEDIC) by 4.97\%. These improvements highlight \sysname{}'s ability to accurately detect OOD instances with fewer false positives.

In terms of ID classification accuracy, \sysname{} performs well across all datasets. It achieves the highest accuracy of 68.60\% on \textsc{ColoredMNIST} and 85.38\% on \textsc{PACS}. While not the top performer on \textsc{VLCS} and \textsc{TerraIncognita}, \sysname{} remains competitive, close to the best baselines.

The consistency of \sysname{}'s performance across different datasets and OOD detection algorithms underscores its generalizability and potential applicability to a wide range of real-world scenarios. Moreover, \sysname{}'s ability to maintain high ID classification accuracy while excelling in OOD detection demonstrates its effectiveness in balancing these two crucial aspects of robust machine learning models.

Interestingly, the results reveal that in our highly imbalanced setting, which features a relatively small proportion of OOD instances, the open-set domain generalization approaches do not consistently outperform traditional domain generalization baselines. This could stem from their overreliance on basic threshold-based OOD identification strategies and their complex network designs, which limit their integration with mainstream OOD detectors. In contrast, \sysname{}'s flexibility in incorporating various OOD detection algorithms allows it to adapt more effectively to different datasets and scenarios.

\begin{table*}[!t]
\centering
\caption{Per-Domain performance comparison for each dataset. The top-performing scores are highlighted in \textbf{bold}, and the second-best scores are \underline{underlined}.}
\label{tab:per-domain-performance}
\resizebox{\textwidth}{!}{%
\begin{tabular}{c lccc|ccc|ccc|ccc}

\toprule

\multirow{2}{*}{Dataset} & \multirow{2}{*}{Methods} & \multicolumn{3}{c}{+90\%} & \multicolumn{3}{c}{+80\%} & \multicolumn{3}{c}{-90\%} & \multicolumn{3}{c}{} \\
\cmidrule(lr){3-5} \cmidrule(lr){6-8} \cmidrule(lr){9-11} \cmidrule(lr){12-14}
 & & \textbf{AUROC} & \textbf{AUPR} & \textbf{Accuracy} & \textbf{AUROC} & \textbf{AUPR} & \textbf{Accuracy} & \textbf{AUROC} & \textbf{AUPR} & \textbf{Accuracy} & \textbf{AUROC} & \textbf{AUPR} & \textbf{Accuracy} \\
\midrule
\multirow{12}{*}{\rotatebox[origin=c]{90}{\textsc{ColoredMNIST}}} 
 & ERM* \cite{vapnik1999nature} & 50.54\phantom{0} $\pm$ \phantom{0}5.6 & 11.47\phantom{0} $\pm$ \phantom{0}0.4 & 43.91\phantom{0} $\pm$ \phantom{0}3.7 & 48.74\phantom{0} $\pm$ \phantom{0}5.8 & 10.49\phantom{0} $\pm$ \phantom{0}0.3 & 49.69\phantom{0} $\pm$ \phantom{0}1.5 & 57.23\phantom{0} $\pm$ \phantom{0}5.3 & 10.21\phantom{0} $\pm$ \phantom{0}0.3 & 10.21\phantom{0} $\pm$ \phantom{0}0.3 & \phantom{00.000} - \phantom{00.00} & \phantom{00.000} - \phantom{00.00} & \phantom{00.000} - \phantom{00.00} \\
 & IRM* \cite{arjovsky2019invariant} & 53.17\phantom{0} $\pm$ \phantom{0}0.5 & 11.27\phantom{0} $\pm$ \phantom{0}0.4 & 51.63\phantom{0} $\pm$ \phantom{0}1.9 & 52.51\phantom{0} $\pm$ \phantom{0}3.0 & 11.23\phantom{0} $\pm$ \phantom{0}1.2 & 59.84\phantom{0} $\pm$ \phantom{0}1.6 & 53.28\phantom{0} $\pm$ \phantom{0}1.2 & 10.94\phantom{0} $\pm$ \phantom{0}0.7 & 12.85\phantom{0} $\pm$ \phantom{0}2.2 & \phantom{00.000} - \phantom{00.00} & \phantom{00.000} - \phantom{00.00} & \phantom{00.000} - \phantom{00.00} \\
 & Mixup* \cite{yan2020improve} & 55.84\phantom{0} $\pm$ \phantom{0}1.9 & 12.08\phantom{0} $\pm$ \phantom{0}1.1 & 50.02\phantom{0} $\pm$ \phantom{0}0.8 & 52.65\phantom{0} $\pm$ \phantom{0}3.1 & 11.23\phantom{0} $\pm$ \phantom{0}1.3 & 48.85\phantom{0} $\pm$ \phantom{0}1.2 & 51.20\phantom{0} $\pm$ \phantom{0}3.7 & 09.52\phantom{0} $\pm$ \phantom{0}0.2 & 10.17\phantom{0} $\pm$ \phantom{0}0.1 & \phantom{00.000} - \phantom{00.00} & \phantom{00.000} - \phantom{00.00} & \phantom{00.000} - \phantom{00.00} \\
 & MBDG* \cite{robey2021model} & 71.54\phantom{0} $\pm$ \phantom{0}9.0 & 18.88\phantom{0} $\pm$ \phantom{0}3.9 & 68.67\phantom{0} $\pm$ \phantom{0}2.8 & 69.70\phantom{0} $\pm$ \phantom{0}4.2 & 16.40\phantom{0} $\pm$ \phantom{0}2.2 & 70.54\phantom{0} $\pm$ \phantom{0}2.9 & 47.46\phantom{0} $\pm$ \phantom{0}6.4 & 10.19\phantom{0} $\pm$ \phantom{0}0.8 & 19.16\phantom{0} $\pm$ \phantom{0}3.3 & \phantom{00.000} - \phantom{00.00} & \phantom{00.000} - \phantom{00.00} & \phantom{00.000} - \phantom{00.00} \\
 & EDir-MMD \cite{noguchi2023simple} & 49.45\phantom{0} $\pm$ 12.5 & 30.53\phantom{0} $\pm$ 19.1 & 73.81\phantom{0} $\pm$ \phantom{0}0.8 & 48.73\phantom{0} $\pm$ 12.2 & 31.35\phantom{0} $\pm$ 19.5 & 72.75\phantom{0} $\pm$ \phantom{0}0.7 & 43.73\phantom{0} $\pm$ 15.4 & 35.93\phantom{0} $\pm$ 29.6 & \underline{35.31\phantom{0}} $\pm$ \phantom{0}5.1 & \phantom{00.000} - \phantom{00.00} & \phantom{00.000} - \phantom{00.00} & \phantom{00.000} - \phantom{00.00} \\
 & EDst-MMD \cite{noguchi2023simple} & 51.98\phantom{0} $\pm$ \phantom{0}8.9 & 32.57\phantom{0} $\pm$ 21.4 & 74.11\phantom{0} $\pm$ \phantom{0}0.1 & 65.15\phantom{0} $\pm$ \phantom{0}0.6 & 40.94\phantom{0} $\pm$ 26.7 & 73.51\phantom{0} $\pm$ \phantom{0}3.5 & 40.14\phantom{0} $\pm$ \phantom{0}7.7 & 30.41\phantom{0} $\pm$ 23.7 & \underline{35.31\phantom{0}} $\pm$ \phantom{0}5.1 & \phantom{00.000} - \phantom{00.00} & \phantom{00.000} - \phantom{00.00} & \phantom{00.000} - \phantom{00.00} \\
 & SCONE \cite{bai2023feed} & 60.45\phantom{0} $\pm$ \phantom{0}3.8 & 23.45\phantom{0} $\pm$ \phantom{0}1.2 & 71.36\phantom{0} $\pm$ \phantom{0}2.0 & 59.12\phantom{0} $\pm$ \phantom{0}4.2 & 22.18\phantom{0} $\pm$ \phantom{0}1.4 & 70.68\phantom{0} $\pm$ \phantom{0}2.3 & 56.44\phantom{0} $\pm$ \phantom{0}4.1 & 20.76\phantom{0} $\pm$ \phantom{0}1.5 & 22.64\phantom{0} $\pm$ \phantom{0}3.0 & \phantom{00.000} - \phantom{00.00} & \phantom{00.000} - \phantom{00.00} & \phantom{00.000} - \phantom{00.00} \\
 & DAML \cite{shu2021open} & 55.71\phantom{0} $\pm$ \phantom{0}0.7 & \underline{38.21\phantom{0}} $\pm$ \phantom{0}0.4 & 74.94\phantom{0} $\pm$ \phantom{0}0.2 & 53.99\phantom{0} $\pm$ \phantom{0}0.3 & 36.87\phantom{0} $\pm$ \phantom{0}0.3 & \underline{75.08\phantom{0}} $\pm$ \phantom{0}0.2 & 45.75\phantom{0} $\pm$ \phantom{0}2.1 & 29.48\phantom{0} $\pm$ \phantom{0}0.9 & 10.31\phantom{0} $\pm$ \phantom{0}0.0 & \phantom{00.000} - \phantom{00.00} & \phantom{00.000} - \phantom{00.00} & \phantom{00.000} - \phantom{00.00} \\
 & MEDIC \cite{wang2023generalizable} & 63.21\phantom{0} $\pm$ \phantom{0}4.3 & \textbf{38.62\phantom{0}} $\pm$ \phantom{0}2.9 & \textbf{75.12\phantom{0}} $\pm$ \phantom{0}0.5 & 61.87\phantom{0} $\pm$ \phantom{0}4.7 & 36.89\phantom{0} $\pm$ \phantom{0}3.1 & \textbf{76.34\phantom{0}} $\pm$ \phantom{0}0.6 & 57.77\phantom{0} $\pm$ \phantom{0}4.8 & 33.42\phantom{0} $\pm$ \phantom{0}3.2 & 29.59\phantom{0} $\pm$ \phantom{0}2.4 & \phantom{00.000} - \phantom{00.00} & \phantom{00.000} - \phantom{00.00} & \phantom{00.000} - \phantom{00.00} \\
 \cmidrule{2-14}
 & \sysname{}-DDU & \textbf{68.73\phantom{0}} $\pm$ \phantom{0}3.1 & 36.51\phantom{0} $\pm$ \phantom{0}0.3 & 72.60\phantom{0} $\pm$ \phantom{0}0.3 & 61.98\phantom{0} $\pm$ \phantom{0}3.4 & 36.05\phantom{0} $\pm$ \phantom{0}0.2 & 72.90\phantom{0} $\pm$ \phantom{0}0.3 & 51.41\phantom{0} $\pm$ \phantom{0}3.3 & \textbf{38.52\phantom{0}} $\pm$ \phantom{0}0.4 & \textbf{60.10\phantom{0}} $\pm$ \phantom{0}2.8 & \phantom{00.000} - \phantom{00.00} & \phantom{00.000} - \phantom{00.00} & \phantom{00.000} - \phantom{00.00} \\
 & \sysname{}-MSP & \underline{63.33\phantom{0}} $\pm$ \phantom{0}6.9 & 37.29\phantom{0} $\pm$ \phantom{0}3.1 & 72.60\phantom{0} $\pm$ \phantom{0}0.3 & \underline{72.40\phantom{0}} $\pm$ \phantom{0}6.1 & \underline{44.38\phantom{0}} $\pm$ \phantom{0}4.9 & 72.90\phantom{0} $\pm$ \phantom{0}0.3 & \textbf{59.35\phantom{0}} $\pm$ \phantom{0}7.8 & \underline{35.84\phantom{0}} $\pm$ \phantom{0}3.2 & \textbf{60.10\phantom{0}} $\pm$ \phantom{0}2.8 & \phantom{00.000} - \phantom{00.00} & \phantom{00.000} - \phantom{00.00} & \phantom{00.000} - \phantom{00.00} \\
 & \sysname{}-Energy & 63.30\phantom{0} $\pm$ \phantom{0}7.0 & 37.37\phantom{0} $\pm$ \phantom{0}3.2 & 72.60\phantom{0} $\pm$ \phantom{0}0.3 & \textbf{72.97\phantom{0}} $\pm$ \phantom{0}6.0 & \textbf{44.99\phantom{0}} $\pm$ \phantom{0}4.7 & 72.90\phantom{0} $\pm$ \phantom{0}0.3 & \underline{59.24\phantom{0}} $\pm$ \phantom{0}7.6 & 35.80\phantom{0} $\pm$ \phantom{0}3.3 & \textbf{60.10\phantom{0}} $\pm$ \phantom{0}2.8 & \phantom{00.000} - \phantom{00.00} & \phantom{00.000} - \phantom{00.00} & \phantom{00.000} - \phantom{00.00} \\
 
\toprule

\multirow{2}{*}{Dataset} & \multirow{2}{*}{Methods} & \multicolumn{3}{c}{P} & \multicolumn{3}{c}{A} & \multicolumn{3}{c}{C} & \multicolumn{3}{c}{S} \\
\cmidrule(lr){3-5} \cmidrule(lr){6-8} \cmidrule(lr){9-11} \cmidrule(lr){12-14}
 & & \textbf{AUROC} & \textbf{AUPR} & \textbf{Accuracy} & \textbf{AUROC} & \textbf{AUPR} & \textbf{Accuracy} & \textbf{AUROC} & \textbf{AUPR} & \textbf{Accuracy} & \textbf{AUROC} & \textbf{AUPR} & \textbf{Accuracy} \\
\midrule
\multirow{12}{*}{\rotatebox[origin=c]{90}{\textsc{PACS}}} 
 & ERM* \cite{vapnik1999nature} & 83.22\phantom{0} $\pm$ \phantom{0}6.7 & 48.01\phantom{0} $\pm$ \phantom{0}7.1 & \underline{96.31\phantom{0}} $\pm$ \phantom{0}0.9 & 80.00\phantom{0} $\pm$ \phantom{0}5.8 & 38.38\phantom{0} $\pm$ 10.7 & \underline{87.92\phantom{0}} $\pm$ \phantom{0}1.6 & 74.90\phantom{0} $\pm$ \phantom{0}4.1 & 31.81\phantom{0} $\pm$ \phantom{0}7.1 & 79.36\phantom{0} $\pm$ \phantom{0}1.7 & 82.38\phantom{0} $\pm$ \phantom{0}5.3 & \underline{37.62\phantom{0}} $\pm$ 14.1 & 74.34\phantom{0} $\pm$ \phantom{0}0.4 \\
 & IRM* \cite{arjovsky2019invariant} & \underline{93.41\phantom{0}} $\pm$ \phantom{0}0.8 & \underline{64.63\phantom{0}} $\pm$ \phantom{0}8.5 & 95.16\phantom{0} $\pm$ \phantom{0}0.6 & 78.05\phantom{0} $\pm$ \phantom{0}7.2 & 36.08\phantom{0} $\pm$ \phantom{0}8.5 & 86.90\phantom{0} $\pm$ \phantom{0}1.3 & 72.18\phantom{0} $\pm$ \phantom{0}9.5 & 30.65\phantom{0} $\pm$ 12.6 & 79.22\phantom{0} $\pm$ \phantom{0}1.8 & 80.64\phantom{0} $\pm$ \phantom{0}2.0 & 37.21\phantom{0} $\pm$ 10.0 & 74.41\phantom{0} $\pm$ \phantom{0}3.1 \\
 & Mixup* \cite{yan2020improve} & 87.13\phantom{0} $\pm$ \phantom{0}1.7 & 54.33\phantom{0} $\pm$ \phantom{0}8.1 & \textbf{96.82\phantom{0}} $\pm$ \phantom{0}0.6 & 75.17\phantom{0} $\pm$ \phantom{0}5.5 & 36.13\phantom{0} $\pm$ \phantom{0}9.5 & \textbf{88.93\phantom{0}} $\pm$ \phantom{0}0.9 & 66.83\phantom{0} $\pm$ \phantom{0}4.8 & 20.71\phantom{0} $\pm$ \phantom{0}6.7 & 79.80\phantom{0} $\pm$ \phantom{0}1.6 & 69.68\phantom{0} $\pm$ \phantom{0}2.7 & 29.70\phantom{0} $\pm$ 10.1 & 74.13\phantom{0} $\pm$ \phantom{0}2.5 \\
 & MBDG* \cite{robey2021model} & 85.37\phantom{0} $\pm$ \phantom{0}5.1 & 47.32\phantom{0} $\pm$ \phantom{0}5.7 & 80.70\phantom{0} $\pm$ 10.9 & 75.62\phantom{0} $\pm$ \phantom{0}4.0 & 29.49\phantom{0} $\pm$ \phantom{0}4.2 & 82.67\phantom{0} $\pm$ \phantom{0}0.9 & 69.28\phantom{0} $\pm$ \phantom{0}5.4 & 26.60\phantom{0} $\pm$ \phantom{0}9.6 & 72.87\phantom{0} $\pm$ \phantom{0}1.0 & \textbf{83.76\phantom{0}} $\pm$ \phantom{0}1.9 & \textbf{41.97\phantom{0}} $\pm$ \phantom{0}9.8 & 81.34\phantom{0} $\pm$ \phantom{0}1.5 \\
 & EDir-MMD \cite{noguchi2023simple} & 79.25\phantom{0} $\pm$ \phantom{0}3.4 & 42.50\phantom{0} $\pm$ 12.0 & 71.92\phantom{0} $\pm$ \phantom{0}0.1 & 80.07\phantom{0} $\pm$ \phantom{0}4.1 & 37.63\phantom{0} $\pm$ \phantom{0}5.5 & 83.23\phantom{0} $\pm$ \phantom{0}3.4 & 70.91\phantom{0} $\pm$ \phantom{0}0.9 & 22.06\phantom{0} $\pm$ 10.1 & 83.18\phantom{0} $\pm$ \phantom{0}3.3 & 79.23\phantom{0} $\pm$ \phantom{0}7.0 & 29.22\phantom{0} $\pm$ 21.3 & \underline{81.89\phantom{0}} $\pm$ \phantom{0}1.5 \\
 & EDst-MMD \cite{noguchi2023simple} & 79.81\phantom{0} $\pm$ \phantom{0}3.8 & 45.89\phantom{0} $\pm$ \phantom{0}8.6 & 75.08\phantom{0} $\pm$ \phantom{0}2.4 & 80.43\phantom{0} $\pm$ \phantom{0}5.0 & 38.41\phantom{0} $\pm$ \phantom{0}3.1 & 82.17\phantom{0} $\pm$ \phantom{0}3.1 & 73.80\phantom{0} $\pm$ \phantom{0}1.2 & 25.18\phantom{0} $\pm$ 10.2 & \textbf{85.29\phantom{0}} $\pm$ \phantom{0}2.5 & 77.17\phantom{0} $\pm$ \phantom{0}2.6 & 21.84\phantom{0} $\pm$ 13.6 & \textbf{83.23\phantom{0}} $\pm$ \phantom{0}2.1 \\
 & SCONE \cite{bai2023feed} & 68.52\phantom{0} $\pm$ \phantom{0}2.8 & 29.13\phantom{0} $\pm$ \phantom{0}7.9 & 84.45\phantom{0} $\pm$ \phantom{0}0.6 & 73.89\phantom{0} $\pm$ \phantom{0}3.1 & 32.76\phantom{0} $\pm$ \phantom{0}9.2 & 81.22\phantom{0} $\pm$ \phantom{0}0.4 & 70.45\phantom{0} $\pm$ \phantom{0}4.2 & 28.91\phantom{0} $\pm$ \phantom{0}8.3 & 80.98\phantom{0} $\pm$ \phantom{0}0.5 & 72.38\phantom{0} $\pm$ \phantom{0}3.9 & 31.76\phantom{0} $\pm$ \phantom{0}8.6 & 81.59\phantom{0} $\pm$ \phantom{0}0.5 \\
 & DAML \cite{shu2021open} & 89.61\phantom{0} $\pm$ \phantom{0}1.0 & 56.89\phantom{0} $\pm$ 15.3 & 90.56\phantom{0} $\pm$ \phantom{0}2.3 & 79.50\phantom{0} $\pm$ \phantom{0}0.3 & 34.53\phantom{0} $\pm$ 11.6 & 82.14\phantom{0} $\pm$ \phantom{0}0.9 & 69.48\phantom{0} $\pm$ \phantom{0}8.1 & 21.79\phantom{0} $\pm$ 14.3 & 77.06\phantom{0} $\pm$ \phantom{0}0.8 & 55.11\phantom{0} $\pm$ \phantom{0}6.0 & 11.98\phantom{0} $\pm$ \phantom{0}8.3 & 73.52\phantom{0} $\pm$ \phantom{0}1.1 \\
 & MEDIC \cite{wang2023generalizable} & 85.62\phantom{0} $\pm$ \phantom{0}2.0 & 58.37\phantom{0} $\pm$ 12.9 & 95.17\phantom{0} $\pm$ \phantom{0}0.2 & 81.31\phantom{0} $\pm$ \phantom{0}1.5 & \underline{38.42\phantom{0}} $\pm$ 13.6 & 83.34\phantom{0} $\pm$ \phantom{0}0.7 & 68.62\phantom{0} $\pm$ 10.1 & 23.74\phantom{0} $\pm$ 16.4 & 77.59\phantom{0} $\pm$ \phantom{0}0.0 & 72.24\phantom{0} $\pm$ \phantom{0}2.9 & 22.86\phantom{0} $\pm$ \phantom{0}1.1 & 78.99\phantom{0} $\pm$ \phantom{0}2.2 \\
 \cmidrule{2-14}
 & \sysname{}-DDU & \textbf{98.12\phantom{0}} $\pm$ \phantom{0}0.0 & \textbf{89.43\phantom{0}} $\pm$ \phantom{0}0.8 & 95.02\phantom{0} $\pm$ \phantom{0}1.2 & \textbf{87.52\phantom{0}} $\pm$ \phantom{0}1.9 & \textbf{45.55\phantom{0}} $\pm$ \phantom{0}6.2 & 86.40\phantom{0} $\pm$ \phantom{0}0.4 & \underline{86.56\phantom{0}} $\pm$ \phantom{0}1.2 & \underline{41.06\phantom{0}} $\pm$ \phantom{0}8.0 & \underline{84.12\phantom{0}} $\pm$ \phantom{0}0.5 & \underline{82.54\phantom{0}} $\pm$ \phantom{0}2.6 & 27.60\phantom{0} $\pm$ 16.7 & 75.97\phantom{0} $\pm$ \phantom{0}1.1 \\
 & \sysname{}-MSP & 90.34\phantom{0} $\pm$ \phantom{0}0.3 & 55.22\phantom{0} $\pm$ \phantom{0}7.2 & 95.02\phantom{0} $\pm$ \phantom{0}1.2 & \underline{80.59\phantom{0}} $\pm$ \phantom{0}2.3 & 36.16\phantom{0} $\pm$ 15.1 & 86.40\phantom{0} $\pm$ \phantom{0}0.4 & 81.28\phantom{0} $\pm$ \phantom{0}2.4 & 29.52\phantom{0} $\pm$ \phantom{0}6.1 & \underline{84.12\phantom{0}} $\pm$ \phantom{0}0.5 & 72.60\phantom{0} $\pm$ \phantom{0}4.0 & 15.58\phantom{0} $\pm$ \phantom{0}6.5 & 75.97\phantom{0} $\pm$ \phantom{0}1.1 \\
 & \sysname{}-Energy & 87.94\phantom{0} $\pm$ \phantom{0}1.6 & 50.98\phantom{0} $\pm$ \phantom{0}0.2 & 95.02\phantom{0} $\pm$ \phantom{0}1.2 & 79.94\phantom{0} $\pm$ \phantom{0}0.0 & 34.49\phantom{0} $\pm$ 11.6 & 86.40\phantom{0} $\pm$ \phantom{0}0.4 & \textbf{86.80\phantom{0}} $\pm$ \phantom{0}6.0 & \textbf{48.18\phantom{0}} $\pm$ \phantom{0}6.0 & \underline{84.12\phantom{0}} $\pm$ \phantom{0}0.5 & 72.45\phantom{0} $\pm$ \phantom{0}4.9 & 16.10\phantom{0} $\pm$ \phantom{0}5.8 & 75.97\phantom{0} $\pm$ \phantom{0}1.1 \\

\toprule

\multirow{2}{*}{Dataset} & \multirow{2}{*}{Methods} & \multicolumn{3}{c}{V} & \multicolumn{3}{c}{L} & \multicolumn{3}{c}{C} & \multicolumn{3}{c}{S} \\
\cmidrule(lr){3-5} \cmidrule(lr){6-8} \cmidrule(lr){9-11} \cmidrule(lr){12-14}
 & & \textbf{AUROC} & \textbf{AUPR} & \textbf{Accuracy} & \textbf{AUROC} & \textbf{AUPR} & \textbf{Accuracy} & \textbf{AUROC} & \textbf{AUPR} & \textbf{Accuracy} & \textbf{AUROC} & \textbf{AUPR} & \textbf{Accuracy} \\
\midrule
\multirow{12}{*}{\rotatebox[origin=c]{90}{\textsc{VLCS}}} 
 & ERM* \cite{vapnik1999nature} & 63.81\phantom{0} $\pm$ \phantom{0}3.7 & 17.41\phantom{0} $\pm$ \phantom{0}3.8 & 81.04\phantom{0} $\pm$ \phantom{0}0.9 & 73.24\phantom{0} $\pm$ \phantom{0}2.6 & 09.27\phantom{0} $\pm$ \phantom{0}0.0 & 64.43\phantom{0} $\pm$ \phantom{0}1.0 & \underline{93.62\phantom{0}} $\pm$ \phantom{0}3.0 & \underline{64.01\phantom{0}} $\pm$ \phantom{0}6.4 & \underline{99.15\phantom{0}} $\pm$ \phantom{0}0.0 & 57.48\phantom{0} $\pm$ \phantom{0}2.0 & 19.82\phantom{0} $\pm$ 19.1 & \underline{75.95\phantom{0}} $\pm$ \phantom{0}0.4 \\
 & IRM* \cite{arjovsky2019invariant} & 63.52\phantom{0} $\pm$ \phantom{0}8.6 & 18.22\phantom{0} $\pm$ \phantom{0}5.6 & 78.20\phantom{0} $\pm$ \phantom{0}2.9 & 73.63\phantom{0} $\pm$ \phantom{0}4.3 & 07.94\phantom{0} $\pm$ \phantom{0}1.7 & 64.57\phantom{0} $\pm$ \phantom{0}1.2 & 90.44\phantom{0} $\pm$ \phantom{0}6.4 & 63.62\phantom{0} $\pm$ \phantom{0}1.4 & 98.26\phantom{0} $\pm$ \phantom{0}0.2 & 62.58\phantom{0} $\pm$ \phantom{0}0.6 & 21.50\phantom{0} $\pm$ 20.2 & 73.08\phantom{0} $\pm$ \phantom{0}3.4 \\
 & Mixup* \cite{yan2020improve} & 58.34\phantom{0} $\pm$ \phantom{0}1.2 & 16.04\phantom{0} $\pm$ \phantom{0}8.8 & 79.21\phantom{0} $\pm$ \phantom{0}2.8 & 67.34\phantom{0} $\pm$ 12.7 & \textbf{31.93\phantom{0}} $\pm$ 25.0 & 65.90\phantom{0} $\pm$ \phantom{0}1.4 & 90.02\phantom{0} $\pm$ \phantom{0}0.4 & 36.34\phantom{0} $\pm$ 20.2 & 98.83\phantom{0} $\pm$ \phantom{0}0.0 & 57.02\phantom{0} $\pm$ \phantom{0}3.9 & 22.20\phantom{0} $\pm$ 15.5 & 75.81\phantom{0} $\pm$ \phantom{0}0.1 \\
 & MBDG* \cite{robey2021model} & 60.78\phantom{0} $\pm$ \phantom{0}5.9 & 14.92\phantom{0} $\pm$ \phantom{0}1.0 & 77.02\phantom{0} $\pm$ \phantom{0}0.0 & 74.20\phantom{0} $\pm$ \phantom{0}4.8 & 07.99\phantom{0} $\pm$ \phantom{0}1.8 & 64.10\phantom{0} $\pm$ \phantom{0}0.4 & 87.46\phantom{0} $\pm$ \phantom{0}5.0 & 42.27\phantom{0} $\pm$ 20.2 & \textbf{99.23\phantom{0}} $\pm$ \phantom{0}0.1 & 65.11\phantom{0} $\pm$ \phantom{0}6.7 & 19.44\phantom{0} $\pm$ 18.0 & 71.02\phantom{0} $\pm$ \phantom{0}0.7 \\
 & EDir-MMD \cite{noguchi2023simple} & 58.93\phantom{0} $\pm$ \phantom{0}6.4 & 13.30\phantom{0} $\pm$ \phantom{0}1.4 & \underline{82.25\phantom{0}} $\pm$ \phantom{0}0.4 & 51.41\phantom{0} $\pm$ 20.1 & 04.99\phantom{0} $\pm$ \phantom{0}2.7 & 63.31\phantom{0} $\pm$ \phantom{0}2.1 & 58.93\phantom{0} $\pm$ \phantom{0}0.5 & 16.53\phantom{0} $\pm$ \phantom{0}6.7 & 71.02\phantom{0} $\pm$ \phantom{0}0.9 & 53.38\phantom{0} $\pm$ \phantom{0}8.8 & 13.90\phantom{0} $\pm$ 13.1 & 74.72\phantom{0} $\pm$ \phantom{0}1.2 \\
 & EDst-MMD \cite{noguchi2023simple} & 57.90\phantom{0} $\pm$ \phantom{0}8.2 & 13.56\phantom{0} $\pm$ \phantom{0}2.4 & \textbf{82.83\phantom{0}} $\pm$ \phantom{0}0.9 & 48.38\phantom{0} $\pm$ 17.5 & 03.47\phantom{0} $\pm$ \phantom{0}1.2 & \underline{68.11\phantom{0}} $\pm$ \phantom{0}2.9 & 50.01\phantom{0} $\pm$ 10.1 & 12.31\phantom{0} $\pm$ \phantom{0}0.9 & 70.43\phantom{0} $\pm$ \phantom{0}0.6 & 56.38\phantom{0} $\pm$ \phantom{0}8.1 & 14.70\phantom{0} $\pm$ 13.9 & 74.88\phantom{0} $\pm$ \phantom{0}1.8 \\
 & SCONE \cite{bai2023feed} & 63.33\phantom{0} $\pm$ \phantom{0}1.0 & \underline{20.94\phantom{0}} $\pm$ \phantom{0}3.6 & 80.18\phantom{0} $\pm$ \phantom{0}2.4 & 63.95\phantom{0} $\pm$ \phantom{0}1.1 & 22.63\phantom{0} $\pm$ \phantom{0}3.7 & \textbf{76.32\phantom{0}} $\pm$ \phantom{0}2.5 & 66.42\phantom{0} $\pm$ \phantom{0}1.3 & 27.85\phantom{0} $\pm$ \phantom{0}3.2 & 79.45\phantom{0} $\pm$ \phantom{0}1.8 & 65.78\phantom{0} $\pm$ \phantom{0}1.4 & 26.18\phantom{0} $\pm$ \phantom{0}3.4 & 75.89\phantom{0} $\pm$ \phantom{0}2.1 \\
 & DAML \cite{shu2021open} & 52.09\phantom{0} $\pm$ \phantom{0}1.0 & 10.75\phantom{0} $\pm$ \phantom{0}0.8 & 74.57\phantom{0} $\pm$ \phantom{0}1.7 & 65.46\phantom{0} $\pm$ \phantom{0}2.2 & 05.02\phantom{0} $\pm$ \phantom{0}0.9 & 65.00\phantom{0} $\pm$ \phantom{0}0.4 & 85.83\phantom{0} $\pm$ \phantom{0}0.5 & 55.95\phantom{0} $\pm$ \phantom{0}0.1 & 97.37\phantom{0} $\pm$ \phantom{0}0.5 & 60.28\phantom{0} $\pm$ \phantom{0}1.7 & 19.65\phantom{0} $\pm$ \phantom{0}8.7 & \textbf{76.86\phantom{0}} $\pm$ \phantom{0}1.6 \\
 & MEDIC \cite{wang2023generalizable} & \underline{64.25\phantom{0}} $\pm$ \phantom{0}4.8 & 21.28\phantom{0} $\pm$ \phantom{0}6.8 & 79.88\phantom{0} $\pm$ \phantom{0}2.4 & \underline{79.50\phantom{0}} $\pm$ \phantom{0}0.2 & 09.19\phantom{0} $\pm$ \phantom{0}0.7 & 63.22\phantom{0} $\pm$ \phantom{0}1.0 & 87.08\phantom{0} $\pm$ \phantom{0}0.7 & \textbf{74.79\phantom{0}} $\pm$ \phantom{0}6.0 & 98.91\phantom{0} $\pm$ \phantom{0}0.5 & 60.50\phantom{0} $\pm$ \phantom{0}2.1 & 20.29\phantom{0} $\pm$ 19.1 & 70.94\phantom{0} $\pm$ \phantom{0}2.6 \\
 \cmidrule{2-14}
 & \sysname{}-DDU & \textbf{66.06\phantom{0}} $\pm$ 17.8 & \textbf{26.40\phantom{0}} $\pm$ 10.4 & 74.88\phantom{0} $\pm$ \phantom{0}4.4 & \textbf{82.53\phantom{0}} $\pm$ \phantom{0}8.2 & \underline{27.97\phantom{0}} $\pm$ 12.4 & 66.56\phantom{0} $\pm$ \phantom{0}0.8 & \textbf{95.14\phantom{0}} $\pm$ \phantom{0}0.3 & 63.75\phantom{0} $\pm$ 11.5 & 97.92\phantom{0} $\pm$ \phantom{0}0.4 & \textbf{75.12\phantom{0}} $\pm$ \phantom{0}8.3 & \underline{27.29\phantom{0}} $\pm$ 17.3 & 74.75\phantom{0} $\pm$ \phantom{0}0.4 \\
 & \sysname{}-MSP & 62.50\phantom{0} $\pm$ \phantom{0}2.8 & 20.14\phantom{0} $\pm$ \phantom{0}2.0 & 74.88\phantom{0} $\pm$ \phantom{0}4.4 & 68.55\phantom{0} $\pm$ \phantom{0}1.7 & 11.28\phantom{0} $\pm$ \phantom{0}0.7 & 66.56\phantom{0} $\pm$ \phantom{0}0.8 & 86.48\phantom{0} $\pm$ \phantom{0}6.3 & 55.54\phantom{0} $\pm$ 24.2 & 97.92\phantom{0} $\pm$ \phantom{0}0.4 & 66.59\phantom{0} $\pm$ \phantom{0}0.0 & 27.10\phantom{0} $\pm$ 20.9 & 74.75\phantom{0} $\pm$ \phantom{0}0.4 \\
 & \sysname{}-Energy & 54.90\phantom{0} $\pm$ \phantom{0}0.8 & 16.70\phantom{0} $\pm$ \phantom{0}0.8 & 74.88\phantom{0} $\pm$ \phantom{0}4.4 & 66.13\phantom{0} $\pm$ \phantom{0}5.6 & 13.58\phantom{0} $\pm$ \phantom{0}1.6 & 66.56\phantom{0} $\pm$ \phantom{0}0.8 & 88.58\phantom{0} $\pm$ \phantom{0}6.4 & 57.99\phantom{0} $\pm$ 22.3 & 97.92\phantom{0} $\pm$ \phantom{0}0.4 & \underline{67.84\phantom{0}} $\pm$ \phantom{0}0.7 & \textbf{29.35\phantom{0}} $\pm$ 21.6 & 74.75\phantom{0} $\pm$ \phantom{0}0.4 \\
 
\toprule

\multirow{2}{*}{Dataset} & \multirow{2}{*}{Methods} & \multicolumn{3}{c}{L38} & \multicolumn{3}{c}{L43} & \multicolumn{3}{c}{L46} & \multicolumn{3}{c}{L100} \\
\cmidrule(lr){3-5} \cmidrule(lr){6-8} \cmidrule(lr){9-11} \cmidrule(lr){12-14}
 & & \textbf{AUROC} & \textbf{AUPR} & \textbf{Accuracy} & \textbf{AUROC} & \textbf{AUPR} & \textbf{Accuracy} & \textbf{AUROC} & \textbf{AUPR} & \textbf{Accuracy} & \textbf{AUROC} & \textbf{AUPR} & \textbf{Accuracy} \\
\midrule
\multirow{12}{*}{\rotatebox[origin=c]{90}{\textsc{TerraIncognita}}} 
 & ERM* \cite{vapnik1999nature} & 48.12\phantom{0} $\pm$ \phantom{0}2.4 & 10.01\phantom{0} $\pm$ \phantom{0}3.9 & 56.70\phantom{0} $\pm$ \phantom{0}5.0 & 50.53\phantom{0} $\pm$ \phantom{0}2.3 & 08.75\phantom{0} $\pm$ \phantom{0}3.3 & 58.60\phantom{0} $\pm$ \phantom{0}3.7 & 53.59\phantom{0} $\pm$ \phantom{0}3.6 & 25.71\phantom{0} $\pm$ \phantom{0}5.8 & 49.20\phantom{0} $\pm$ \phantom{0}1.1 & 50.01\phantom{0} $\pm$ \phantom{0}7.5 & 20.47\phantom{0} $\pm$ \phantom{0}8.0 & 44.20\phantom{0} $\pm$ \phantom{0}2.7 \\
 & IRM* \cite{arjovsky2019invariant} & 54.49\phantom{0} $\pm$ \phantom{0}4.9 & 10.99\phantom{0} $\pm$ \phantom{0}5.8 & 45.00\phantom{0} $\pm$ 13.7 & 52.94\phantom{0} $\pm$ \phantom{0}2.8 & 05.90\phantom{0} $\pm$ \phantom{0}2.9 & \underline{63.20\phantom{0}} $\pm$ \phantom{0}2.2 & 57.00\phantom{0} $\pm$ \phantom{0}4.1 & \textbf{27.21\phantom{0}} $\pm$ \phantom{0}7.8 & 47.20\phantom{0} $\pm$ \phantom{0}2.3 & 52.28\phantom{0} $\pm$ 10.0 & 27.40\phantom{0} $\pm$ 10.7 & \underline{52.50\phantom{0}} $\pm$ \phantom{0}9.8 \\
 & Mixup* \cite{yan2020improve} & 53.04\phantom{0} $\pm$ \phantom{0}4.9 & 10.46\phantom{0} $\pm$ \phantom{0}5.4 & 53.10\phantom{0} $\pm$ \phantom{0}0.1 & 53.62\phantom{0} $\pm$ \phantom{0}2.6 & 05.23\phantom{0} $\pm$ \phantom{0}2.1 & \textbf{63.50\phantom{0}} $\pm$ \phantom{0}0.6 & 53.16\phantom{0} $\pm$ \phantom{0}2.8 & 20.56\phantom{0} $\pm$ \phantom{0}5.0 & 43.50\phantom{0} $\pm$ \phantom{0}4.0 & \textbf{60.17\phantom{0}} $\pm$ \phantom{0}9.1 & \textbf{31.09\phantom{0}} $\pm$ 10.8 & \textbf{57.50\phantom{0}} $\pm$ \phantom{0}1.8 \\
 & MBDG* \cite{robey2021model} & 55.04\phantom{0} $\pm$ \phantom{0}6.6 & 12.79\phantom{0} $\pm$ \phantom{0}5.5 & 36.80\phantom{0} $\pm$ 13.3 & 54.78\phantom{0} $\pm$ \phantom{0}3.4 & 07.13\phantom{0} $\pm$ \phantom{0}2.4 & 56.20\phantom{0} $\pm$ \phantom{0}0.9 & 52.32\phantom{0} $\pm$ \phantom{0}4.3 & 20.19\phantom{0} $\pm$ \phantom{0}4.6 & 41.20\phantom{0} $\pm$ \phantom{0}1.1 & 49.01\phantom{0} $\pm$ \phantom{0}7.4 & 27.13\phantom{0} $\pm$ 11.9 & 41.70\phantom{0} $\pm$ \phantom{0}7.3 \\
 & EDir-MMD \cite{noguchi2023simple} & 61.43\phantom{0} $\pm$ 18.3 & \textbf{19.73\phantom{0}} $\pm$ 18.0 & 54.78\phantom{0} $\pm$ \phantom{0}0.1 & 38.80\phantom{0} $\pm$ 13.9 & 04.66\phantom{0} $\pm$ \phantom{0}2.5 & 60.41\phantom{0} $\pm$ \phantom{0}1.1 & 59.11\phantom{0} $\pm$ \phantom{0}9.0 & 23.87\phantom{0} $\pm$ \phantom{0}5.8 & \textbf{57.80\phantom{0}} $\pm$ \phantom{0}1.1 & \underline{58.21\phantom{0}} $\pm$ 18.1 & 25.73\phantom{0} $\pm$ 21.5 & 40.34\phantom{0} $\pm$ \phantom{0}7.3 \\
 & EDst-MMD \cite{noguchi2023simple} & 60.01\phantom{0} $\pm$ 19.0 & \underline{19.54\phantom{0}} $\pm$ 18.0 & \underline{57.99\phantom{0}} $\pm$ \phantom{0}1.5 & 42.13\phantom{0} $\pm$ 11.8 & 05.00\phantom{0} $\pm$ \phantom{0}2.6 & 57.99\phantom{0} $\pm$ \phantom{0}3.1 & \underline{59.68\phantom{0}} $\pm$ \phantom{0}7.1 & 25.66\phantom{0} $\pm$ 10.3 & \underline{57.06\phantom{0}} $\pm$ \phantom{0}0.5 & 54.41\phantom{0} $\pm$ 19.5 & 23.86\phantom{0} $\pm$ 19.8 & 42.91\phantom{0} $\pm$ \phantom{0}4.9 \\
 & SCONE \cite{bai2023feed} & 55.82\phantom{0} $\pm$ \phantom{0}9.1 & 17.62\phantom{0} $\pm$ \phantom{0}4.2 & 51.20\phantom{0} $\pm$ \phantom{0}2.9 & 46.25\phantom{0} $\pm$ \phantom{0}8.2 & 12.85\phantom{0} $\pm$ \phantom{0}3.8 & 54.63\phantom{0} $\pm$ \phantom{0}3.6 & 48.98\phantom{0} $\pm$ \phantom{0}8.7 & 15.78\phantom{0} $\pm$ \phantom{0}4.3 & 49.26\phantom{0} $\pm$ \phantom{0}2.9 & 53.47\phantom{0} $\pm$ \phantom{0}8.9 & 18.95\phantom{0} $\pm$ \phantom{0}4.5 & 48.75\phantom{0} $\pm$ \phantom{0}3.4 \\
 & DAML \cite{shu2021open} & 62.98\phantom{0} $\pm$ 10.3 & 14.71\phantom{0} $\pm$ 12.8 & 48.30\phantom{0} $\pm$ \phantom{0}7.5 & 54.59\phantom{0} $\pm$ \phantom{0}7.2 & 06.38\phantom{0} $\pm$ \phantom{0}3.3 & 49.80\phantom{0} $\pm$ \phantom{0}1.5 & 53.52\phantom{0} $\pm$ 10.7 & 19.24\phantom{0} $\pm$ \phantom{0}2.8 & 42.20\phantom{0} $\pm$ \phantom{0}1.3 & 52.35\phantom{0} $\pm$ 11.4 & 24.36\phantom{0} $\pm$ 21.5 & 49.40\phantom{0} $\pm$ \phantom{0}0.1 \\
 & MEDIC \cite{wang2023generalizable} & 64.90\phantom{0} $\pm$ \phantom{0}5.6 & 14.05\phantom{0} $\pm$ 11.5 & 43.92\phantom{0} $\pm$ \phantom{0}6.5 & 50.74\phantom{0} $\pm$ \phantom{0}5.6 & 05.89\phantom{0} $\pm$ \phantom{0}3.4 & 50.02\phantom{0} $\pm$ \phantom{0}0.4 & 56.68\phantom{0} $\pm$ 10.0 & 20.46\phantom{0} $\pm$ \phantom{0}3.2 & 43.13\phantom{0} $\pm$ \phantom{0}2.8 & 54.35\phantom{0} $\pm$ \phantom{0}4.8 & 27.42\phantom{0} $\pm$ 24.2 & 50.12\phantom{0} $\pm$ \phantom{0}5.9 \\
 \cmidrule{2-14}
 & \sysname{}-DDU & \underline{65.32\phantom{0}} $\pm$ 11.3 & 15.66\phantom{0} $\pm$ 13.7 & \textbf{61.16\phantom{0}} $\pm$ \phantom{0}1.1 & \textbf{62.54\phantom{0}} $\pm$ \phantom{0}2.2 & \underline{12.90\phantom{0}} $\pm$ \phantom{0}0.5 & 59.99\phantom{0} $\pm$ \phantom{0}4.2 & 58.18\phantom{0} $\pm$ \phantom{0}9.1 & \underline{26.12\phantom{0}} $\pm$ 17.0 & 46.09\phantom{0} $\pm$ \phantom{0}6.5 & 56.78\phantom{0} $\pm$ \phantom{0}7.8 & 26.44\phantom{0} $\pm$ 23.4 & 44.98\phantom{0} $\pm$ \phantom{0}6.7 \\
 & \sysname{}-MSP & \textbf{68.37\phantom{0}} $\pm$ \phantom{0}0.3 & 12.80\phantom{0} $\pm$ \phantom{0}9.7 & \textbf{61.16\phantom{0}} $\pm$ \phantom{0}1.1 & \underline{57.82\phantom{0}} $\pm$ \phantom{0}0.5 & 07.72\phantom{0} $\pm$ \phantom{0}5.6 & 59.99\phantom{0} $\pm$ \phantom{0}4.2 & 59.08\phantom{0} $\pm$ \phantom{0}2.9 & 22.56\phantom{0} $\pm$ 11.6 & 46.09\phantom{0} $\pm$ \phantom{0}6.5 & 56.30\phantom{0} $\pm$ \phantom{0}3.2 & \underline{30.60\phantom{0}} $\pm$ 28.6 & 44.98\phantom{0} $\pm$ \phantom{0}6.7 \\
 & \sysname{}-Energy & 58.56\phantom{0} $\pm$ \phantom{0}2.9 & 12.70\phantom{0} $\pm$ 10.3 & \textbf{61.16\phantom{0}} $\pm$ \phantom{0}1.1 & 54.88\phantom{0} $\pm$ \phantom{0}8.3 & \textbf{16.53\phantom{0}} $\pm$ \phantom{0}6.8 & 59.99\phantom{0} $\pm$ \phantom{0}4.2 & \textbf{62.54\phantom{0}} $\pm$ \phantom{0}2.9 & 25.73\phantom{0} $\pm$ 14.0 & 46.09\phantom{0} $\pm$ \phantom{0}6.5 & 55.35\phantom{0} $\pm$ \phantom{0}2.5 & 27.22\phantom{0} $\pm$ 25.0 & 44.98\phantom{0} $\pm$ \phantom{0}6.7 \\

 \bottomrule

\end{tabular}%
}
\\[3pt] % Adds vertical space after the table
\raggedright % Aligns text to the left
\textit{* The reported values for these baselines represent their best performance across all three OOD detectors: DDU, MSP, and Energy.}
\end{table*}

Table \ref{tab:per-domain-performance} presents the performance breakdown across different domains in the \textsc{ColoredMNIST}, \textsc{PACS}, \textsc{VLCS}, and \textsc{TerraIncognita} datasets. The \sysname{}-* methods consistently demonstrate strong performance across all datasets, often outperforming other baselines. In \textsc{ColoredMNIST}, \sysname{} variants achieve significantly higher AUROC and AUPR scores, particularly in the challenging ``-90\%" domain. For \textsc{PACS}, \sysname{}-DDU excels in most domains, notably achieving 98.12\% AUROC and 89.43\% AUPR in the ``Photo (P)" domain. In \textsc{VLCS}, \sysname{} methods show remarkable improvements, with \sysname{}-DDU achieving the highest AUROC scores across all domains. The \textsc{TerraIncognita} dataset proves challenging for all methods, but \sysname{} variants still manage to outperform baselines in several instances, particularly in AUROC scores. Interestingly, while \sysname{}-DDU often leads in performance, \sysname{}-MSP and \sysname{}-Energy show competitive results in specific scenarios, highlighting the adaptability of the \sysname{} approach. Despite the varied nature of domain shifts across these datasets, \sysname{} methods consistently maintain competitive in-domain classification accuracy while significantly improving OOD detection, demonstrating a robust and balanced performance across diverse domain generalization tasks.

\begin{figure}[!t]
\centering
    \begin{minipage}[b]{0.49\linewidth}
       \includegraphics[width=\linewidth]{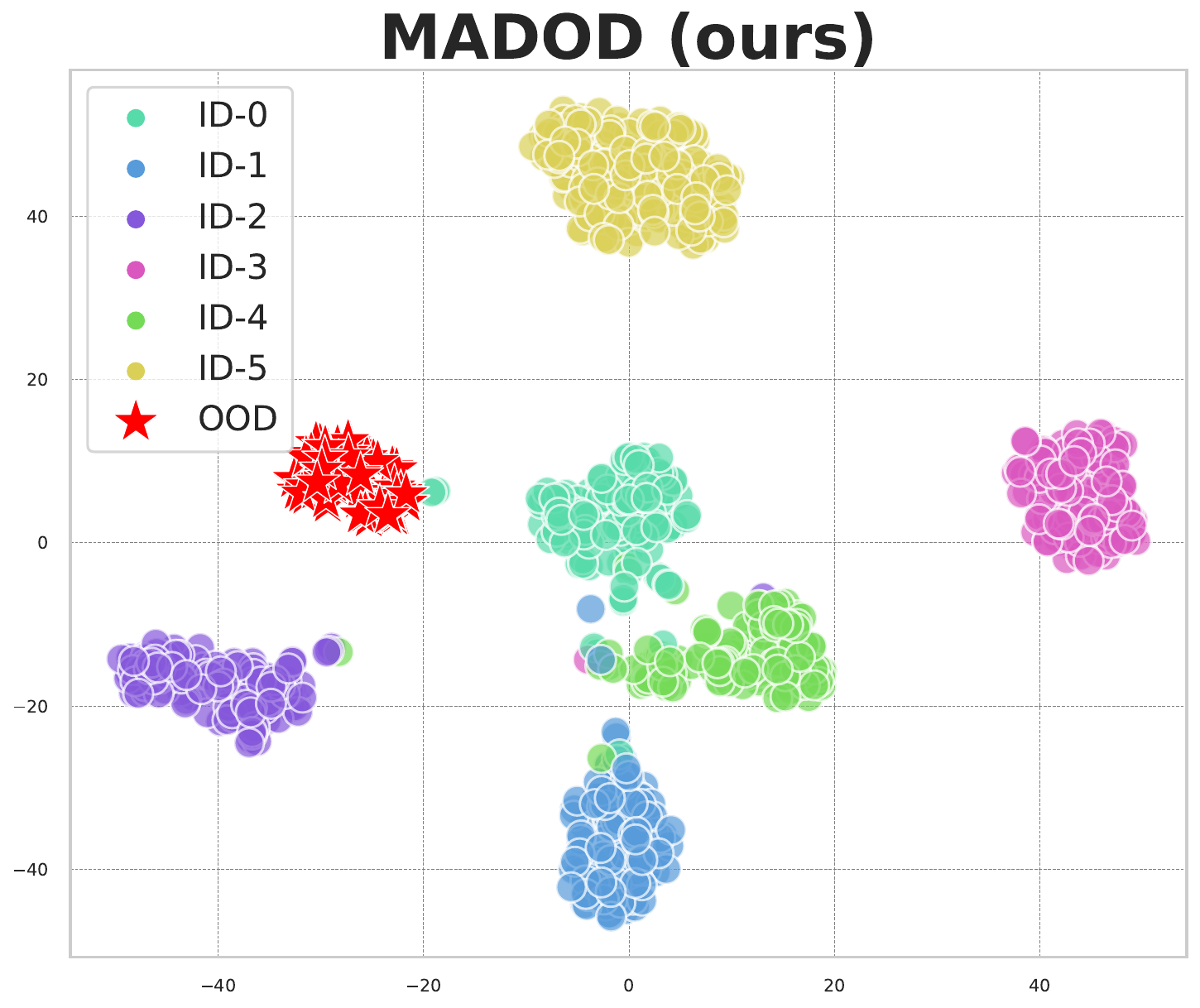}
       % \caption{t-SNE visualization of \sysname{}}
       \label{fig:Ng2}
    \end{minipage}
    \begin{minipage}[b]{0.49\linewidth}
       \includegraphics[width=\linewidth]{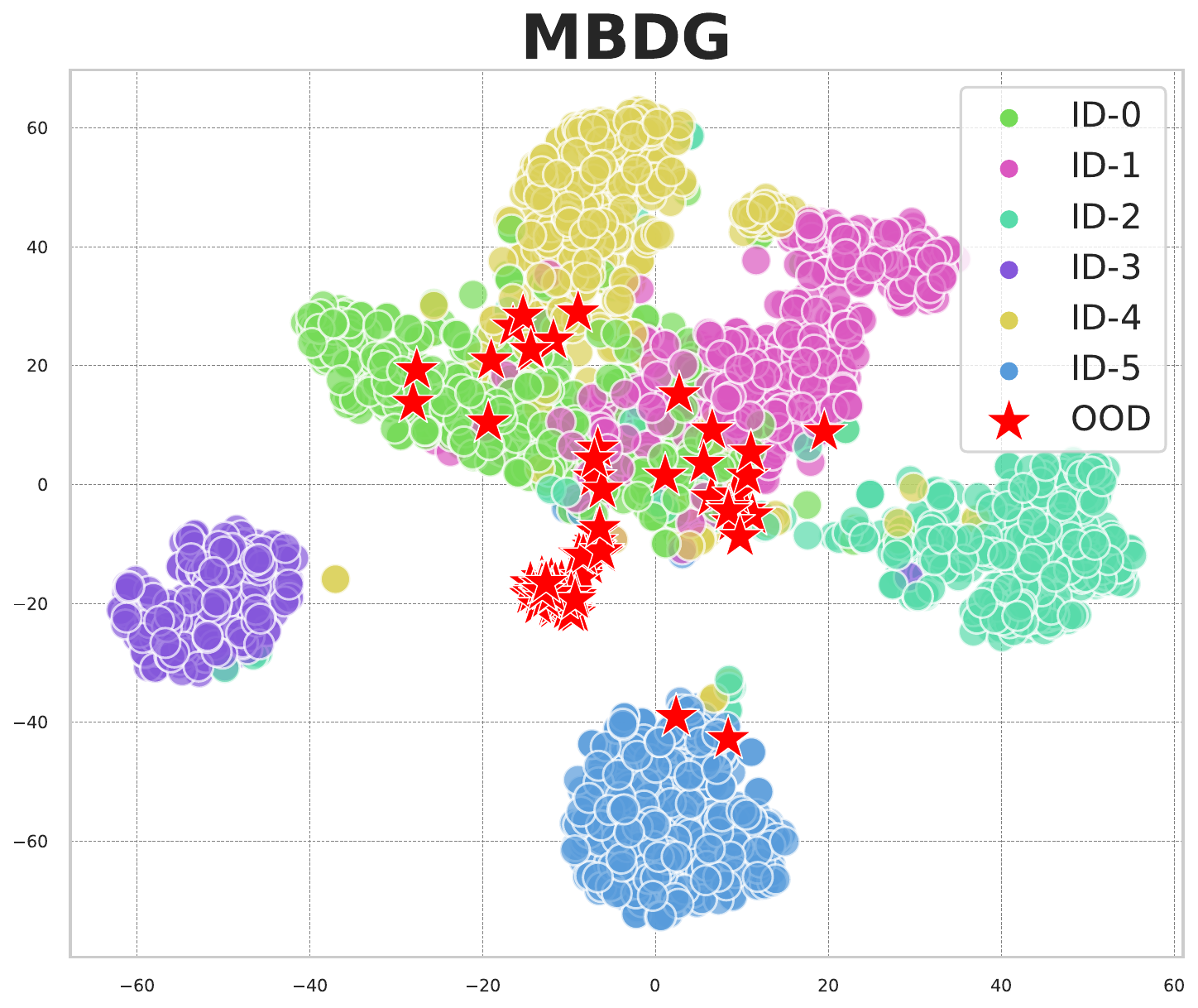}
       % \caption{t-SNE visualization of MBDG}
       \label{fig:Ng1}
    \end{minipage}%
    \vspace{-10mm}
    \caption{Comparison of t-SNE visualizations of the latent feature space for \sysname{} (ours) versus MBDG. These plots are generated on the \textsc{PACS} dataset, with ``Photo" as the test domain. Red stars are OODs of the same semantic label.}
\label{fig:tsne-compare}
\end{figure}

\textbf{t-SNE Visualization Comparison.} Figure \ref{fig:tsne-compare} presents t-SNE visualizations of the latent feature space for MBDG (a) and \sysname{} (b), revealing crucial differences in their data representations. \sysname{}'s plot demonstrates clearer boundaries between ID classes and tighter intra-class clustering compared to MBDG, suggesting more discriminative feature learning. Notably, the OOD class (class 6, pink) in \sysname{}'s visualization shows significantly improved separation from ID classes with minimal overlap, a key improvement over MBDG's representation. This clear OOD-ID separation in \sysname{}'s latent space directly aligns with its superior OOD detection performance. These characteristics of \sysname{}'s latent space, particularly the enhanced OOD-ID separability, provide strong visual evidence for its improved performance in both OOD detection and domain generalization tasks.

\begin{figure}[!t]
    \centering
    \includegraphics[width=\linewidth]{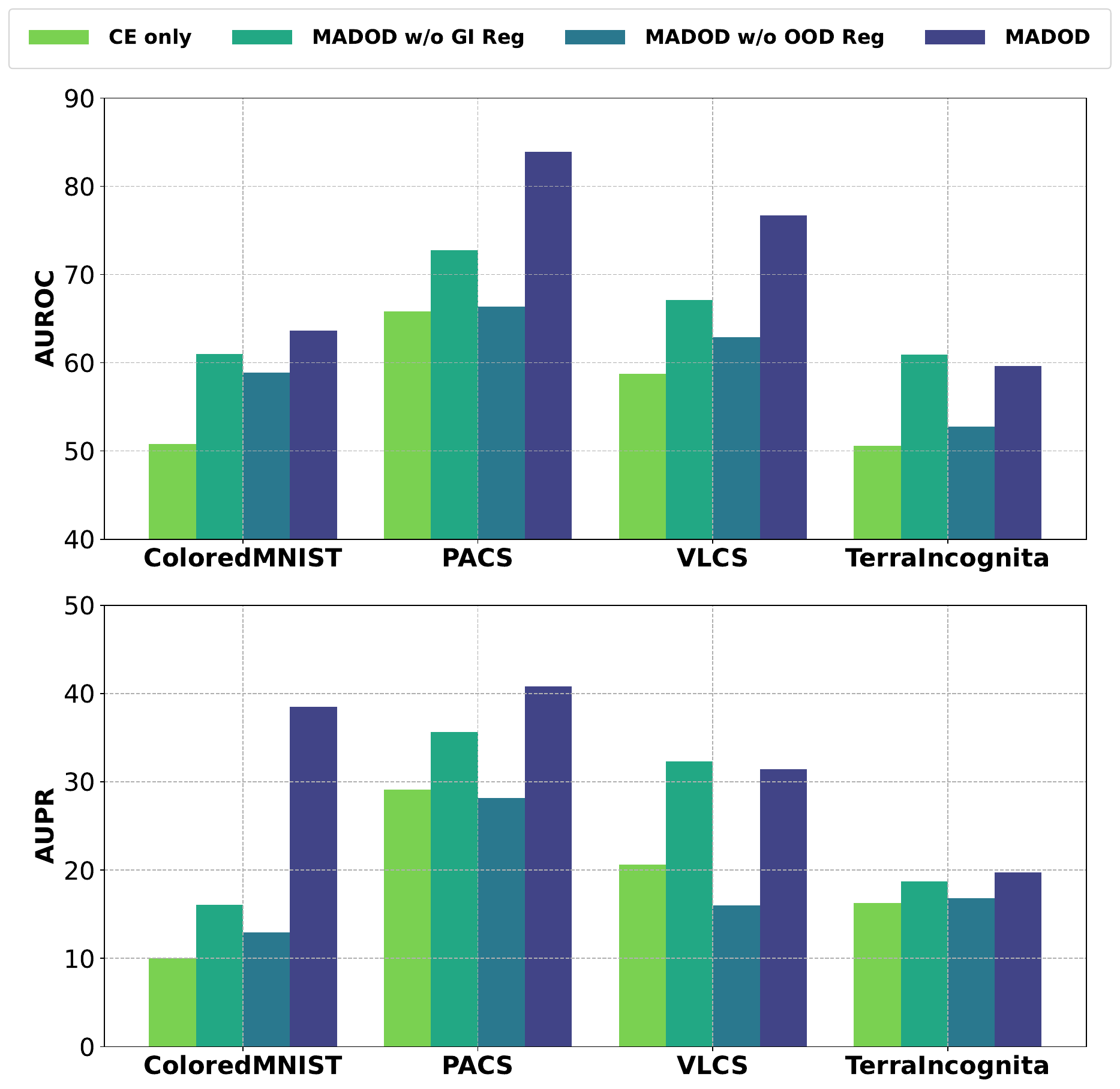}
    \vspace{-5mm}
    \caption{Ablation study results in terms of AUROC and AUPR, comparing: 'CE only' (baseline with cross-entropy), 'MADOD w/o GI Reg' (without GI regularization), 'MADOD w/o OOD Reg' (without OOD regularization), and 'MADOD' (full model with both regularizations).}
    \label{fig:ablation-study}
\end{figure}

\textbf{Ablation Studies.} We conducted an ablation study to evaluate the impact of $G$-invariance (GI) and OOD regularization in \sysname{}. Figure \ref{fig:ablation-study} presents results across \textsc{ColoredMNIST}, \textsc{PACS}, \textsc{VLCS}, and \textsc{TerraIncognita} datasets.

The full \sysname{} framework consistently outperforms ablated versions across all datasets. On \textsc{PACS}, \sysname{} achieves 83.89\% AUROC and 40.82\% AUPR, significantly surpassing ablated variants. Removing GI regularization notably decreases AUROC scores, particularly evident in \textsc{PACS} (83.89\% to 72.74\%). OOD regularization's absence primarily affects AUPR scores, as seen in \textsc{ColoredMNIST} (38.53\% to 12.92\%). The baseline model (CE only) performs worst across all datasets, emphasizing both components' necessity.

This study confirms the substantial contribution of both GI and OOD regularization to \sysname{}'s performance, demonstrating its effectiveness as an integrated system for OOD detection in unseen domains.

\section{Conclusion}
    We present \sysname{} (Meta-learned Across Domain Out-of-distribution Detection), a novel framework for semantic OOD detection in unseen domains. By integrating meta-learning with $G$-invariance and innovative task construction, \sysname{} addresses both covariate shifts and semantic shifts without requiring external OOD training data. Our framework's compatibility with existing OOD detection techniques enhances its versatility for real-world applications.

Comprehensive evaluations on benchmark datasets demonstrate \sysname{}'s superiority over baselines in OOD detection across unseen domains while maintaining competitive ID classification accuracy. These results validate \sysname{}'s effectiveness in generalizing to unknown test environments and identifying semantic OODs, paving the way for more robust and adaptable machine learning systems.

\bibliographystyle{bibtex/IEEEtran}
\bibliography{bibtex/IEEEabrv,bibtex/references}

\end{document}